\pdfoutput=1
\documentclass[11pt]{article}

\usepackage[]{acl}

\usepackage{times}
\usepackage{latexsym}

\usepackage[T1]{fontenc}
\usepackage[utf8]{inputenc}

\usepackage{microtype}
\usepackage{mathtools}  
\usepackage{subcaption}
\usepackage{xcolor,colortbl}
\usepackage{booktabs}
\usepackage{algorithm}
\usepackage{algorithmic}
\usepackage[utf8]{inputenc}
\usepackage{epsfig}
\usepackage{graphicx}
\usepackage{amssymb}
\usepackage{epsfig}
\usepackage{amsmath}
\usepackage{multirow}
\usepackage{makecell}
\DeclareMathOperator*{\argmax}{arg\,max}
\newcolumntype{P}[1]{>{\centering\arraybackslash}p{#1}}
\usepackage{color,soulutf8}

\title{``Diversity and Uncertainty in Moderation'' are the Key to Data Selection for Multilingual Few-shot Transfer}

\author{ Shanu Kumar\textsuperscript{1}  \quad Sandipan Dandapat\textsuperscript{1} \quad Monojit Choudhury\textsuperscript{2}\\
\textsuperscript{1} Microsoft R\&D, Hyderabad, India \\
\textsuperscript{2} Microsoft Research, India   \\
{\tt \small \{shankum,sadandap,monojitc\}@microsoft.com}
}

\begin{document}

\maketitle
\begin{abstract}
%Transfer learning using pretrained multilingual transformer-based language models is extensively used for many NLP tasks with good zero- and few-shot cross-lingual transfer. 
Few-shot transfer often shows substantial gain over zero-shot transfer~\cite{lauscher2020zero}, which is a practically useful trade-off between fully supervised and unsupervised learning approaches for multilingual pretrained model-based systems. 
This paper explores various strategies for selecting data for annotation that can result in a better few-shot transfer. The proposed approaches rely on multiple measures such as data entropy using $n$-gram language model, predictive entropy, and gradient embedding. 
% The data cross-entropy helps diversity sampling; predictive entropy uses the fine-tuned model's task-specific learning, while the gradient embedding method integrates diversity and uncertainty. 
We propose a loss embedding method for sequence labeling tasks, which induces diversity and uncertainty sampling similar to gradient embedding. The proposed data selection strategies are evaluated and compared for POS tagging, NER, and NLI tasks for up to 20 languages. Our experiments show that the gradient and loss embedding-based strategies consistently outperform random data selection baselines, with gains varying with the initial performance of the zero-shot transfer. Furthermore, the proposed method shows similar trends in improvement even when the model is fine-tuned using a lower proportion of the original task-specific labeled training data for zero-shot transfer.
\end{abstract}

\section{Introduction}
Language resource distribution, for both labeled and unlabeled data, across the world's languages is extremely skewed, with more than 95\% of the languages having hardly any task-specific labeled data~\cite{joshi-etal-2020-state}. Therefore, \textit{cross-lingual zero-shot transfer} using pretrained deep multilingual language models has received significant attention from the NLP community. During cross-lingual zero-shot transfer, first a multilingual model \cite{devlin-etal-2019-bert,NEURIPS2019_c04c19c2,conneau2020unsupervised,liu2020multilingual,xue2020mt5,ouyang2020ernie} is created using only unlabelled data from a large number of languages (typically in the range of 100) with some self-supervised learning objectives. These pretrained models are then fine-tuned with task-specific labeled data from one or more languages (we refer to these as the \textit{pivot} languages) and tested on all the other languages (here referred to as the \textit{target} languages) for which no annotated data was used during fine-tuning.   
%This observation reflects in the recent success in building deep multilingual language models have led to widespread use of cross-lingual transfer for NLP tasks. 

Many recent work \cite{pires-etal-2019-multilingual, karthikeyan2019cross, wu-dredze-2019-beto, artetxe-etal-2020-cross, lauscher2020zero} have studied the efficacy of zero-shot cross-lingual transfer across languages and factors influencing it. Other work have shown that a few-shot transfer, where very little labeled data in the {\em target} language is also used during fine-tuning,  can result in substantial gains over the zero-shot transfer.  For instance, \citet{lauscher2020zero} show that zero-shot transfer does not hold much promise for transfer across typologically different languages or when there is not enough unlabeled data in the target language during model pretraining. In such cases, the gap in the cross-lingual transfer can be effectively reduced by fine-tuning it on a little annotated data in the target language. However, very few languages have readily available annotated resources for different NLP tasks, and collecting annotated data for a large set of target languages can be expensive and time-consuming \cite{dandapat-etal-2009-complex, 10.1145/2362456.2362479, fort2016collaborative}. Therefore, it is essential to carefully select and annotate target language data for a few-shot transfer, reducing the transfer gap effectively.  

Training data selection has been investigated for several NLP tasks, especially for domain adaptation ~\cite{blitzer2007biographies,sogaard2011data,liu2019reinforced}. The majority of these approaches use different techniques to rank the entire data and use top $n$ data points to train the system~\cite{moore-lewis-2010-intelligent}. 
% Some of the work ~\cite{liu2019reinforced} seamlessly integrates data selection with the training process.
In addition, active learning \cite{fu2013survey, settles2008analysis} has been widely used to improve annotation efficiently by using model predictions to select informative data. Active learning is generally used in an iterative setting, in which a model is learned at each iteration, and samples are selected for labeling to improve performance. However, in this paper, we are trying to select a few samples. Hence we are limiting the training to one iteration. In the past,~\citet{chaudhary2019little} have used active learning to annotate only uncertain entity spans for Dutch and Hindi languages. However, to the best of our knowledge, none of these approaches have been studied for a large set of languages in a cross-lingual few-shot transfer setting.

The central goal of this work is to propose specific strategies for data selection (and subsequent annotation) for few-shot learning so that the performance in a target language is maximized, given a data budget.
The main contributions of this work are:
[1] We propose different data selection strategies based on the notions of cross-entropy, predictive entropy, gradient embedding and loss embedding, and perform various reliability analyses of these strategies.
[2] We conduct experiments on a set of 20 typologically diverse languages including some syntactically divergent from the pivot language – English \& Chinese.
[3] We propose a loss embedding-based method for sequence labeling tasks which incorporates both diversity and uncertainty sampling.
[4] Through experiments on three NLP tasks, we show that embedding-based strategies perform consistently better than random data selection baselines, with gains varying with the initial performance of the zero-shot transfer. We also observe several language and data-size dependent trends in the performance across different data selection strategies.
[5] Finally, we provide a concrete set of recommendations for data selection based on features such as zero-shot performance and the amount of unlabeled data available for a target-language.
    
    % \item We perform various reliability analyses of the proposed predictive entropy and uncertainty metrics.

The rest of the paper is organized as follows. The next section introduces the novel data sampling strategies. Section 3, 4, and 5 present the experimental setup, results and related research in the area, respectively. Concluding remarks are made in Section 6.

\section{Data Sampling Strategies}
Assuming we have a pre-trained multilingual language model and enough labelled data in a particular language such as English ({\small EN}) for fine-tuning on a task. We can measure the zero-shot performance on a set of target languages. We observe that zero-shot performances are not uniform and often vary with the typological similarity between the target and pivot language as stated by \cite{pires-etal-2019-multilingual, lauscher2020zero}. Nevertheless, all the target languages show a drop in zero-shot performance compared to the performance achieved in  the pivot language. Hence, there is a cross-lingual transfer gap for all the target languages. This gap can be attributed to the inherent linguistic property of the target languages however, \citet{lauscher2020zero} have shown that the cross-lingual transfer gap can be reduced by fine-tuning on a little annotated data in a target language. 

Consequently, few-shot performance can reduce the transfer gap for all the languages. Given a fixed budget, let say $k$ examples, we want to maximize the few-shot performance in a target language by carefully choosing the effective $k$ examples. To this end, we are proposing several data selection strategies in this section. We compare them with random sampling where $k$ target language examples are randomly selected from task-specific fine-tuning data collection. Note that the sampling strategies are oblivious to the actual labels of the data points, as annotation would follow the data selection step in practice.

%We compare several measures for data selection for few-shot cross-lingual transfer with random sampling where $k$ target language examples are randomly selected from task-specific training data.

\subsection{Data Cross-Entropy (DCE)}
Cross-entropy \cite{moore-lewis-2010-intelligent, axelrod-etal-2011-domain, dara2014active} has been widely used for domain adaptation by selecting in-domain data from a large non-domain-specific (contains both in- and out-domain data) corpus. In our scenario, like \citet{dara2014active}, the target language labeled data acts as the large non-domain-specific corpus and using cross-entropy, novel and diverse data is selected from it. Assuming that there is little overlap between the tokens of the pivot and target language during the zero-shot cross-lingual transfer, we presume that {\em no} in-domain labeled data for a particular target language is available initially. We further assume that we have access to a non-domain-specific collection of data points, the entire target language unlabeled corpus (may or may not be distinct from the pretraining corpus). 

%We want to select the data with low data entropy and high diversity in terms of vocabulary.
First, two N-gram language models $\mathbf{M_I}$ and $\mathbf{M_O}$ are trained on the sentences selected $L_I$ (initially an \textit{empty set}) and sentences left in the target language corpus $L_O$ (staring with the entire corpus), respectively. We use SRILM\footnote{\url{http://www.speech.sri.com/projects/srilm/}}~\cite{conf/interspeech/Stolcke02} to build the N-gram (for N=3) language models. We do not want to select sentences which are similar to already picked $L_I$; hence we measure data cross entropy ($\text{DCE}$) and select sentences from $L_O$ that have high entropy with respect to $L_I$ and low entropy with respect to $L_O$. 
\vspace{-0.5em}
\begin{gather}
        H_{I}(x) = H(\mathbf{M_I}(x)) \\ 
    H_{O}(x) = H(\mathbf{M_O}(x)) \\ 
        \text{DCE}(x) = \frac{H_{O}(x)}{\sum_{s \in L_O}{H_{O}(s)}} - \frac{H_{I}(x)}{\sum_{s \in L_O}{H_{I}(s)}} \label{eqn:de} 
\end{gather}
where $H(x)$ is the measure entropy of a sentence $x$ using a N-gram language model. The size of $L_O$ and $L_I$ will vary across the iterations, therefore we appropriately normalize the entropy $H_{I}$ and $H_{O}$ for calculating cross-entropy.
% where $\text{PP}_{in}(x)$ is the perplexity using the language model $M_I$ trained on the sentences already selected $L_I$, whereas $\text{PP}_{out}(x)$ is the perplexity for the language model $M_O$ trained on the sentences left in the target language data $L_O$.

\begin{algorithm}[!]
\caption{Sentence Selection using DCE}
\label{algo}
\textbf{Input}: \text{Target Language Corpus $D^t$, $g$, $k$}\\
\vspace{-1em}
\begin{algorithmic}[1] %[1] enables line numbers
\STATE $L_I \leftarrow \{\} $, $L_O \leftarrow D^t $
\WHILE{$\text{size}(L_I) < k \text{ AND } L_O \ne \phi$}
\STATE $\mathbf{M_I} \leftarrow \text{TrainLM}(L_I)$
\STATE $\mathbf{M_O} \leftarrow \text{TrainLM}(L_O)$
\FOR{each $s \in L_O$}
    \STATE $H_I(s) \leftarrow H(\mathbf{M_I}(s))$
    \STATE $H_O(s) \leftarrow H(\mathbf{M_O}(s))$
\ENDFOR

\FOR{each $s \in L_O$}
    \STATE Calculate DCE($s$)
\ENDFOR
\STATE $L_g \leftarrow \text{Select top $g$ sentences ranked by DCE}(\cdot)$
\STATE $L_I \leftarrow L_I \cup L_g $
\STATE $L_O \leftarrow L_O - L_g  $
\ENDWHILE
\end{algorithmic}
\end{algorithm}

Algorithm \ref{algo} describes the data selection method using data cross-entropy, where $g$ is the number of data points to be selected in one iteration, and $k$ is the total number of sentences to be selected. The overall time complexity of this method is $\mathcal{O}(nk/g)$, where $n = |D^t|$. For reducing the computation time, we can increase $g$, which we set to 10 in our experiments.

\subsection{Predictive Entropy (PE)}
We employ predictive entropy to measure the task-specific knowledge of a fine-tuned model. For a sequence labelling task, we define the predictive entropy $E(x_i)$ of a token $x_i$ of a sentence $x$ given a fine-tuned model $\mathbf{M}$ as follows:
\vspace{-0.5em}
\begin{gather}
    p({y}_{i}|x_i) = {\mathbf{M}}({x_{i}}) 
\end{gather}
\vspace{-2em}
\begin{gather}
    E(x_i) = - \sum_{j=1}^C p({y}_{i}=c_j|x_i)*\log p({y}_{i}=c_j|x_i)
    \vspace{-1em}
\end{gather}
where $c_1, c_2, \dots c_C$ are the class labels. 
%where $N_x$ is the number of tokens. 

% Mean of all the token's entropy can be used to define predictive entropy of a sentence. However, the mean entropy may vary significantly with the length of the sentence and the individual entropy of each token. Thus, it may have higher values for shorter and much lower values for longer sentences. In contrast, sum of all the token's entropy will have higher values for very long sentences. For penalizing the length, 

We define the predictive entropy of the sentence using the equation \eqref{eqn:pe}:
% Defining predictive entropy of a whole sentence is crucial as the mean of all entropy of tokens in the sentence
% $\text{PE}(x)$ represents the predictive entropy of the whole sentence $x$, 
\vspace{-0.5em}
\begin{gather}
% \text{PE}(x) = (N_x)^{\gamma-1}{\sum_{i=1}^{N_x} E(x_i)} ,\ \;\;\; \gamma \in [0, 1] \label{eqn:pe}
\text{PE}(x) = \frac{1}{N_x}{\sum_{i=1}^{N_x} E(x_i)} \label{eqn:pe}
\vspace{-0.5em}
\end{gather}

% where $\gamma$ controls the importance given to the length of the sentence, $ \gamma=0$ will result in the mean of the tokens' entropy, whereas $\gamma=1$ will be equivalent to sum of the tokens' entropy.
where $N_x$ is the number of tokens in sentence $x$. For classification tasks, $N_x$ will be 1. 

 To define the scoring function for data selection using predictive entropy that can generalize to the corpus with different domain-shift, we use the statistics of the predictive entropy from the entire target language corpus. We use $\mu_{PE}$ mean and $\sigma_{PE}$ standard deviation of the predictive entropy of all the sentences in the corpus. Selecting sentences with very low predictive entropy will not help improve the performance as they have less novel information to enhance the knowledge of the model. Furthermore, picking sentences with very high predictive entropy can be harmful to training. It can be high due to either noise or out-of-domain data. As we want to select very few data instances for few-shot learning and improve further upon the zero-shot performance, we consider selection around $\mu_{PE}$, the mean of the predictive entropy. But if the zero-shot performance is excellent, then $\mu_{PE}$ will be very low, and selecting data closer to mean may not improve over the zero-shot performance. Therefore, we add $\sigma_{PE}$.  
%herefore, adding the $\sigma_{PE}$ standard deviation of predictive entropy to the $\mu_{PE}$ mean will fix this issue. 
We formally define the scoring function in equation \eqref{eqn:pe_algo}.
\vspace{-0.5em}
\begin{gather}
    \text{score}_{PE}(x) = |\text{PE}(x) - (\mu_{PE} + \lambda*\sigma_{PE})| \;\; \label{eqn:pe_algo}
\end{gather}
Here, $\lambda$ controls the distance of the preferred selection zone from $\mu_{PE}$.

{
\begin{table*}[ht]
\scriptsize
\centering
\setlength\tabcolsep{2.5pt}%
%\vspace{-1em}
\begin{tabular}{*{200}{c}}
\toprule
&  &  & \textbf{AR} & \textbf{BG} & \textbf{DE} & \textbf{EL} & \textbf{ES} & \textbf{EU} & \textbf{FI} & \textbf{FR} & \textbf{HE} & \textbf{HI} & \textbf{JA} & \textbf{KO} & \textbf{RU} & \textbf{SV} & \textbf{SW} & \textbf{TH} & \textbf{TR} & \textbf{UR} & \textbf{VI} & \textbf{ZH} \\ 
\textbf{Task} &  \textbf{Model} & \multirow{-2}{*}{\textbf{EN}} & $\triangle$ & $\triangle$ & $\triangle$ & $\triangle$ & $\triangle$ & $\triangle$ & $\triangle$ & $\triangle$ & $\triangle$ & $\triangle$ & $\triangle$ & $\triangle$ & $\triangle$ & $\triangle$ & $\triangle$ & $\triangle$ & $\triangle$ & $\triangle$ & $\triangle$ & $\triangle$\\ 
\midrule
& B & 96.4 & \cellcolor[gray]{0.88}-55.4
 & \cellcolor[gray]{0.95}-11.4 & \cellcolor[gray]{0.95}-11.7 & \cellcolor[gray]{0.95}-18.9 & \cellcolor[gray]{0.95}-13.27 & \cellcolor[gray]{0.88}-37.4 & \cellcolor[gray]{0.95}-19.8 & - & \cellcolor[gray]{0.88}-46.9 & \cellcolor[gray]{0.88}-35.2 & \cellcolor[gray]{0.88}-53.0 & \cellcolor[gray]{0.88}-49.7 & \cellcolor[gray]{0.95}-12.2 & \cellcolor[gray]{0.95}-7.3 & - & - & \cellcolor[gray]{0.95}-26.3 & \cellcolor[gray]{0.88}-43.4 & - & \cellcolor[gray]{0.88}-38.5 \\
 \multirow{-2}{*}{POS} & X & 97.2 & \cellcolor[gray]{0.88}-43.6
 & \cellcolor[gray]{0.95}-9.7 & \cellcolor[gray]{0.95}-9.9 & \cellcolor[gray]{0.95}-14.6 & \cellcolor[gray]{0.95}-13.1 & \cellcolor[gray]{0.88}-27.9 & \cellcolor[gray]{0.95}-14.7 & - & \cellcolor[gray]{0.88}-44.4 & \cellcolor[gray]{0.88}-27.6 & \cellcolor[gray]{0.88}-74.3 & \cellcolor[gray]{0.88}-46.7 & \cellcolor[gray]{0.95}-10.2 & \cellcolor[gray]{0.95}-6.3 & - & - & \cellcolor[gray]{0.95}-20.4 & \cellcolor[gray]{0.88}-37.2 & - & \cellcolor[gray]{0.88}-63.9 \\
\midrule
& B & 84.2 & \cellcolor[gray]{0.88}-45.3 & \cellcolor[gray]{0.95}-7.4
& \cellcolor[gray]{0.95}-6.7 &
\cellcolor[gray]{0.95}-12.8 &
\cellcolor[gray]{0.95}-11.9 & 
\cellcolor[gray]{0.95}-24.3 &
\cellcolor[gray]{0.95}-7.7 & 
\cellcolor[gray]{0.95}-5.7 &
\cellcolor[gray]{0.88}-28.6 & \cellcolor[gray]{0.95}-20.2 & \cellcolor[gray]{0.88}-54.9 & \cellcolor[gray]{0.88}-25.2 & \cellcolor[gray]{0.95}-21.3 & \cellcolor[gray]{0.95}-9.9 & - & \cellcolor[gray]{0.65}-83.5 & \cellcolor[gray]{0.95}-12.4 & \cellcolor[gray]{0.88}-47.9 & \cellcolor[gray]{0.95}-11.3 & \cellcolor[gray]{0.88}-40.7 \\ 
 \multirow{-2}{*}{NER} & X & 82.5 & \cellcolor[gray]{0.88}-39.6 & \cellcolor[gray]{0.95}-5.7 & \cellcolor[gray]{0.95}-9.2 & \cellcolor[gray]{0.95}-9.8 & \cellcolor[gray]{0.95}-8.8 &  \cellcolor[gray]{0.95}-23.9 & \cellcolor[gray]{0.95}--8.7 & \cellcolor[gray]{0.95}-5.6 & \cellcolor[gray]{0.88}-32.5 & \cellcolor[gray]{0.95}-15.3 & \cellcolor[gray]{0.88}-60.5 & \cellcolor[gray]{0.88}-36.6 & \cellcolor[gray]{0.95}-18.7 & \cellcolor[gray]{0.95}-13.4 & - & \cellcolor[gray]{0.65}-78.1 & \cellcolor[gray]{0.95}-8.6 & \cellcolor[gray]{0.88}-34.5 & \cellcolor[gray]{0.95}-16.4 & \cellcolor[gray]{0.88}-53.5 \\ 
 \midrule
 & B &  81.9 & \cellcolor[gray]{0.95}-16.7 & \cellcolor[gray]{0.95}-13.2 & \cellcolor[gray]{0.95}-11.1 & \cellcolor[gray]{0.95}-14.8 & \cellcolor[gray]{0.95}-7.1 & - & - & \cellcolor[gray]{0.95}-7.8 & -  & \cellcolor[gray]{0.88}-22.1 & - & - & \cellcolor[gray]{0.95}-12.9 & - & \cellcolor[gray]{0.88}-32.2 & \cellcolor[gray]{0.88}-28.8 & \cellcolor[gray]{0.95}-20.8 & \cellcolor[gray]{0.88}-24.3 & \cellcolor[gray]{0.95}-11.7 & \cellcolor[gray]{0.95}-13.2 \\ 
 \multirow{-2}{*}{XNLI} & X & 84.1 & \cellcolor[gray]{0.95}-12.7 & \cellcolor[gray]{0.95}-6.3 & \cellcolor[gray]{0.95}-8.3 & \cellcolor[gray]{0.95}-8.8 & \cellcolor[gray]{0.95}-5.7 & - & - & \cellcolor[gray]{0.95}-6.5 & -  & \cellcolor[gray]{0.95}-15.0 & - & - & \cellcolor[gray]{0.95}-9.0 & - & \cellcolor[gray]{0.95}-20.4 & \cellcolor[gray]{0.95}-12.7 & \cellcolor[gray]{0.95}-12.0 & \cellcolor[gray]{0.95}-18.6 & \cellcolor[gray]{0.95}-9.9 & \cellcolor[gray]{0.95}-10.6 \\
%   \midrule
%   & B &  75.3 & \cellcolor[gray]{0.95}-13.9 & \cellcolor[gray]{0.95}-10.1 & \cellcolor[gray]{0.95}-7.9 & \cellcolor[gray]{0.95}-12.2 & \cellcolor[gray]{0.95}-4.9 & - & - & \cellcolor[gray]{0.95}-6.2 & -  & \cellcolor[gray]{0.88}-18.2 & - & - & \cellcolor[gray]{0.95}-10.9 & - & \cellcolor[gray]{0.88}-25.4 & \cellcolor[gray]{0.88}-25.0 & \cellcolor[gray]{0.95}-15.2 & \cellcolor[gray]{0.88}-19.3 & \cellcolor[gray]{0.95}-9.9 & \cellcolor[gray]{0.95}-10.2 \\ 
%  \multirow{-2}{*}{XNLI} & X & 79.0 & \cellcolor[gray]{0.95}-11.1 & \cellcolor[gray]{0.95}-5.8 & \cellcolor[gray]{0.95}-8.3 & \cellcolor[gray]{0.95}-8.8 & \cellcolor[gray]{0.95}-5.7 & - & - & \cellcolor[gray]{0.95}-6.5 & -  & \cellcolor[gray]{0.95}-15.0 & - & - & \cellcolor[gray]{0.95}-9.0 & - & \cellcolor[gray]{0.95}-20.4 & \cellcolor[gray]{0.95}-12.7 & \cellcolor[gray]{0.95}-12.0 & \cellcolor[gray]{0.95}-18.6 & \cellcolor[gray]{0.95}-9.9 & \cellcolor[gray]{0.95}-10.6 \\
\bottomrule
\end{tabular}
\vspace{-0.5em}
\caption {We report the zero-shot cross-lingual transfer performance drops relative to {\small EN} for all the languages on POS, NER, and XNLI tasks with mBERT (\textbf{B}) and XLM-R (\textbf{X}). The results are medians over three RAND initialization (seeds). The darkness of the cell indicates the drops in zero-shot performance.}
\label{tbl:zero_shot}
\vspace{-0.5em}
\end{table*}}

\subsection{Gradient Embedding (GE)}
Most of the data selection strategies use either representative sampling such as DCE or uncertainty sampling such as PE. Recently, \citet{ash2019deep} proposed {\small BADGE} that combines both diversity and uncertainty sampling. {\small BADGE} uses gradient embedding to capture uncertainty from the model, assuming the norm of the gradients will be smaller if the model is highly certain about its predictions and vice versa. As we don't have access to the ground truth labels, the gradient embedding $g_{x_i}$ $\in \mathbb{R}^d$ is computed for a input sentence $x_i$ by taking model's ($M$) prediction as the true label $\hat{y}_i$.
\vspace{-0.5em}
\begin{gather}
    \hat{y}_i = \argmax \mathbf{M}({x_{i}}) \\
    {g}_{x_i} = \frac{\partial }{\partial \theta_{out}} l_{CE}(\mathbf{M}({x_{i}}), \hat{y}_i) \label{eqn:ge}
\end{gather}
% \vspace{-0.2em}
where $l_{CE}$ is the cross-entropy loss function, $\theta_{out}$ $\in \mathbb{R}^d$ refers to the parameters of last layer and $d$ is the number of parameters. We have used hidden states of the [{\small CLS}] token from last layer classification tasks, hence we have computed the gradients with $\theta_{out}$ as the last layer of the pre-trained models.

{\small BADGE} selects samples by applying $k$-{\small MEANS}++ \cite{arthur2006k} clustering on the gradient embedding. The selection is made on the assumption that examples with gradient embedding of small magnitude will tend to cluster together and not be selected repeatedly. $k$-{\small MEANS}++ tend to select samples that are diverse and highly uncertain. For simplicity, we will call {\small BADGE} method as GE.

As we want to select very few data instances for few-shot learning and improve further upon the zero-shot performance, we consider applying GE selection on examples satisfying the following criteria:
\vspace{-0.5em}
\begin{gather}
  \text{GE} (\mathbf{\lambda}) = \text{GE} (\left\{x \colon g_x > \mu_{g} + \lambda * \sigma_{g}\right\}) \label{eqn:ge_lmd}
\end{gather}
% \vspace{-0.2em}
where $\mu_{g}$ and $\sigma_{g}$ are the mean and standard deviation of magnitude of the gradient embedding of all the examples in the corpus. $\lambda$ controls the final value of the selection criteria.

We noticed that in certain cases selecting samples sharing similar context but having different true labels may be more helpful for few-shot learning. To incorporate this, we propose \textbf{GE($\gamma$)}, which adds $\gamma$ similar examples for each $k$ sample selected using the GE method. As gradient embedding loses information about the sentence, we use Multilingual Sentence XLM-R \cite{reimers2020making} for calculating similarity based on sentences. We do not apply any constraints to ensure similar examples have different true labels but the gradient embedding can be used for ensuring it.

\subsection{Loss Embedding (LE)}
Sequence labelling tasks require prediction over all the tokens of a sentence, and therefore we have to calculate the gradient embedding for each token classification. Considering the maximum number of allowed tokens in a sentence to be $m$, the resulting gradient embedding $g_{x_i}$ will of dimension $d \times {m}$. Due to its high dimensionality, applying $k$-{\small MEANS}++ will be expensive. We solve this dimensionality issue by proposing the Loss Embedding method, which has a dimension of $m$, considering $lm$ is usually less than $d$.

Instead of calculating gradient, we consider only using classification loss at each token. For a sentence $x_i$, we compute loss embedding ${l}_{x_i}$ $\in \mathbb{R}^{m}$ by computing cross-entropy loss for each token by taking the model's prediction as actual labels. As the norm of loss embedding will be smaller if the model is highly certain about its predictions and vice-versa, it satisfies the primary assumption of {\small BADGE} method. Another property preferable for sequential tasks is that the sentences with similar syntax will have a similar structure in the loss embedding as it depends upon the position of tokens in a sentence. Therefore applying $k$-{\small MEANS}++ clustering on the loss embedding will induce both diversity and uncertainty sampling. 

Similar to GE, we also experiment with selection of examples satisfying the following criteria: 
\vspace{-0.5em}
\begin{gather}
  \text{LE} (\lambda) = \text{LE} (\left\{x \colon l_x > \mu_{l} + \lambda * \sigma_{l}\right\}) \label{eqn:le_lmd}
\end{gather}
\vspace{-0.1em}
where $\mu_{l}$ and $\sigma_{l}$ are the mean and standard deviation of magnitude of the loss embedding of all the examples in the corpus.

\section{Experimental Setup}
We conduct various experiments to evaluate effectiveness of our proposed data sampling techniques in a few-shot transfer setting with up to 20 languages from various language families on two different sequential tasks and one classification task.

\subsection{Datasets}
We evaluate our methods on three benchmarks datasets on POS-tagging, NER, and NLI. The complete statistics of training and test data available in each language is provided in Appendix \ref{sec:data}.
\vspace{1em}

\hspace{-1em}\textit{Named Entity Recognition} (NER). We perform NER experiments using NER Wikiann dataset~\cite{rahimi-etal-2019-massively} on 20 languages. We also remove duplicates data points from the training corpus as these will hinder data selection. 

\vspace{1em}
\hspace{-1em}\textit{Part-of-speech Tagging} (POS). We perform POS experiments using Universal Dependency treebanks \cite{nivre-etal-2016-universal} on the same set of languages of NER except French ({\small FR}), Thai ({\small TH}), and, Vietnamese ({\small VI}) due to unavailability of substantial amount of training data after removing duplicates. 

\vspace{1em}
\hspace{-1em}\textit{Cross-lingual Natural Language Inference} (XNLI). The XNLI dataset \cite{conneau2018xnli} consists of translated train, dev and test sets in 14 languages of English hypothesis-premise pairs.
\begin{table*}[ht]
\centering
\small
\setlength\tabcolsep{4pt}%
\begin{tabular}{c|c|ccc|ccc|ccc|ccc|ccc}
\toprule
% & $k$  & 10  & 50 & 100 & 500 & 1000 \\
\multicolumn{1}{c}{} & & \multicolumn{3}{c}{$k=10$}  & \multicolumn{3}{|c}{$k=50$} & \multicolumn{3}{|c}{$k=100$} & \multicolumn{3}{|c}{$k=500$} & \multicolumn{3}{|c}{$k=1000$} \\
    \cmidrule(lr){3-5}\cmidrule(lr){6-8}\cmidrule(lr){9-11}  \cmidrule(lr){12-14} \cmidrule(lr){15-17} 
\multicolumn{1}{c}{} & \multirow{-1}{*}{\textbf{Method}} & $\triangle^{C_1}$   & $\triangle^{C_2}$ &  $\triangle^{C_3}$ & $\triangle^{C_1}$  & $\triangle^{C_2}$ &  $\triangle^{C_3}$ & $\triangle^{C_1}$  & $\triangle^{C_2}$ &  $\triangle^{C_3}$ & $\triangle^{C_1}$  & $\triangle^{C_2}$ &  $\triangle^{C_3}$ & $\triangle^{C_1}$  & $\triangle^{C_2}$ &  $\triangle^{C_3}$ \\ 
\midrule
& RAND & 2.9 & 9.9 & 0.5 &
6.4 & 15.8 & 1.3 &
7.7 & 17.4 & 1.3 &
12.1 & 26.9 & 18.6 &
14.0 & 30.4 & \textbf{31.2}
\\ 
& DCE & 1.8 & 8.4 & {4.0} &
5.1 & 12.8 & {2.8} &
5.2 & 11.8 & {3.3} &
9.9 & 23.0 & 18.8 &
12.3 & 28.5 & 29.2
\\
 & PE ($\lambda = 1$) & 3.1 & 10.0 & \textbf{5.7} &
6.8 & 14.5 & \textbf{4.7} &
7.5 & 16.1 & \textbf{3.9} &
12.4 & 24.5 & \textbf{19.8} &
14.2 & 27.4 & 27.4
\\
& LE & 5.5 & \textbf{11.3} & 2.5 &
7.4 & \textbf{18.4} & 1.1 &
\textbf{8.9} & \textbf{18.9} & 0.7 &
\textbf{13.0} & \textbf{27.6} & {18.9} &
\textbf{14.9} & \textbf{30.6} & 31.0
\\
\multirow{-5}{*}{\rotatebox[origin=c]{90}{mBERT}} & LE ($\lambda=0$) & \textbf{5.6} & 11.0 & 0.7 &
\textbf{8.4} & 18.3 & 0.1 &
8.7 & 18.4 & -0.0 &
12.9 & 26.9 & 15.1 &
14.8 & 30.0 & 29.8
\\
\midrule
& RAND & 1.4 & 8.0 & 0.3 &
6.7 & 15.3 & 0.4 &
7.8 & 16.8 & 1.5 &
12.8 & \textbf{26.1} & \textbf{20.7} &
\textbf{14.6} & {29.4} & {27.7}
\\ 
& DCE & -3.8 & 0.5 & 2.4 &
3.4 & 10.0 & 0.9 &
4.3 & 10.5 & -0.2 &
10.5 & 23.8 & 19.2 &
13.2 & 27.8 & 26.3
\\
 & PE ($\lambda = 1$) & \textbf{4.4} & \textbf{8.6} & {3.6} &
7.0 & \textbf{16.5} & 1.2 &
7.9 & \textbf{17.8} & 0.5 &
12.3 & \textbf{26.1} & 19.0 &
14.2 & \textbf{29.9} & \textbf{28.4} 
\\
& LE & {3.0} & 8.1 & \textbf{5.7} &
\textbf{7.9} & 15.7 & 3.4 &
\textbf{9.0} & 16.7 & 2.4 &
\textbf{13.0} & \textbf{26.1} & 18.2 &
14.5 & 29.1 & 23.4
\\
\multirow{-5}{*}{\rotatebox[origin=c]{90}{XLM-R}} & LE ($\lambda=0$) & 2.4 & {8.2} & 1.4 &
7.4 & {16.0} & \textbf{5.0} &
8.5 & {16.9} & \textbf{4.0} &
\textbf{13.0} & \textbf{26.1} & 16.5 &
14.5 & 29.3 & 23.2
\\
\bottomrule
\end{tabular}
% \vspace{-0.5em}
\caption {Few-shot cross-lingual transfer performance on NER tasks with varying number of target language examples $k$ using {\small EN} as the pivot language. We have reported the $\triangle$ delta between few-shot and zero-shot performance averaged across the languages in each category $C_1$, $C_2$, and $C_3$.} 
\label{tbl:few_shot_ner_en}
% \vspace{-0.5em}
\end{table*}

\begin{table}[ht]
\centering
\small
\setlength\tabcolsep{3pt}%
\begin{tabular}{c|c|cc|cc|cc}
\toprule
% & $k$  & 10  & 50 & 100 & 500 & 1000 \\
\multicolumn{1}{c}{} & & \multicolumn{2}{c}{$k=10$}  & \multicolumn{2}{|c}{$k=50$} & \multicolumn{2}{|c}{$k=100$} 
% & \multicolumn{2}{|c}{$k=500$}
\\
    \cmidrule(lr){3-4}\cmidrule(lr){5-6}\cmidrule(lr){7-8} 
    % \cmidrule(lr){9-10}
\multicolumn{1}{c}{} & \multirow{-1}{*}{\textbf{Method}} & $\triangle^{C_1}$   & $\triangle^{C_2}$ & $\triangle^{C_1}$  & $\triangle^{C_2}$ & $\triangle^{C_1}$  & $\triangle^{C_2}$  
% & $\triangle^{C_1}$  & $\triangle^{C_2}$ 
\\ 
\midrule
 & Rand & 4.1 & 22.6 &
6.7 & 27.5 &
7.3 & 28.0
% 11.9 & 46.5
% 12.5 & 48.4
\\
 & DCE & 2.3 & 18.7 &
5.2 & 24.3 &
6.0 & 25.9
% 11.8 & 46.1
% 12.4 & 48.3
\\
 & PE ($\lambda = 1$) & 4.4 & \textbf{23.4} &
\textbf{7.0} & 27.8 &
7.4 & 28.1
\\
& LE & 3.9 & 20.1 &
6.3 & 26.3 &
7.1 & 26.9
% \textbf{12.1} & 46.9
% \textbf{12.8} & \textbf{48.7}
\\
& LE ($\lambda=0$) & \textbf{4.5} & 21.9 &
{6.8} & 27.3 &
\textbf{7.5} & 27.9
% \textbf{12.1} & \textbf{47.0}
% 12.0 & \textbf{48.7}
\\
\multirow{-5}{*}{\rotatebox[origin=c]{90}{mBERT}} & LE ($\lambda=0.5$) & 4.1 & {23.3} &
{6.8} & \textbf{28.2} &
\textbf{7.5} & \textbf{28.6}
% 11.3 & \textbf{47.0}
% 11.5 & 48.8
\\
\midrule
 & RAND & 3.1 & 24.6 &
5.2 & 28.5 &
5.6 & 28.8
% 9.5 & 42.3
% 9.8 & 44.6
\\
 & DCE & 1.8 & 22.1 &
3.7 & 26.0 &
4.5 & 27.2
% 9.4 & 42.0
% 9.8 & 44.5
\\
 & PE ($\lambda = 1$) & 3.4 & 24.7 &
5.6 & 29.1 &
6.1 & \textbf{29.2}
\\
& LE & 2.9 & 22.1 &
5.5 & 27.6 &
6.2 & 28.5
% \textbf{9.6 } & 42.6
% \textbf{9.9} & \textbf{44.9}
\\
& LE ($\lambda=0$) & 3.1 & 24.9 &
5.6 & 28.6 &
6.1 & 28.8
% 9.5 & 42.7
% 9.3 & \textbf{44.9}
\\
\multirow{-5}{*}{\rotatebox[origin=c]{90}{XLM-R}} & LE ($\lambda=0.5$) & \textbf{3.5} & \textbf{25.2} &
\textbf{5.8} & \textbf{29.2} &
\textbf{6.4} & \textbf{29.2}
% 9.0 & \textbf{42.8}
% 8.9 & \textbf{44.9}
\\
\bottomrule
\end{tabular}
% \vspace{-0.5em}
\caption {Few-shot cross-lingual transfer performance on POS tasks with varying number of target language examples $k$ using {\small EN} as the pivot language. We have reported the $\triangle$ delta between few-shot and zero-shot performance averaged across the languages in each category $C_1$ and $C_2$.}
\label{tbl:few_shot_pos_en}
\vspace{-0.5em}
\end{table}

\subsection{Training Details}
We conduct all our experiment using the  12 layer multilingual mBERT \textit{Base cased} \cite{devlin-etal-2019-bert} and XLM-R \cite{conneau2020unsupervised}. We use the standard fine-tuning technique as described in \cite{devlin-etal-2019-bert, pires-etal-2019-multilingual} for all the experiments.
%We employ 12 layer mBERT \textit{Base cased} \cite{devlin-etal-2019-bert} for both tasks. 
We limit the sentence length to 128 subword tokens and set the batch size as 32. Following \cite{lauscher2020zero}, we fix the number of training epochs to 20 and the learning rate as $2.{10}^{-5}$ for NER and POS. For XNLI, we set the training epochs to 3 for zero-shot and 1 for few-shot training, and learning rate as $3.{10}^{-5}$.  We report $F_1$-score for NER and POS, and accuracy for XNLI. All the reported results are medians over three random initializations (seeds). 

\subsection{Zero-Shot Transfer}
Throughout our experiments, we assume {\small EN} as the pivot language. We report the zero-shot cross-lingual transfer results in Table \ref{tbl:zero_shot}. We observe similar trends in zero-shot performance as reported in \cite{lauscher2020zero}, where there are significant drops in performance for {\small TH, JA, AR, ZH, UR, KO, VI}.
% The zero-shot performance can be seen as the task-specific domain shift from the pivot language {\small EN}, where target languages like ({\small ES, DE, FR}) have lower domain shift. In contrast, languages like ({\small JA, AR, ZH}) have higher domain shifts. 
In {\small TH}, we observe the highest transfer gap with nearly 0 $\text{F}_1$-score, which indicates no cross-lingual transfer has happened.

%  \begin{figure*}[h]
% \begin{subfigure}[t]{0.32\textwidth}
%   \centering
%   \captionsetup{justification=centering}
%     \includegraphics[scale=0.165]{fig/wiki_c1_tk.png}
%   \caption{$C_1$ Target Language Group}
%   \label{fig:wiki_c1}
% \end{subfigure}
%     \hfill
% \begin{subfigure}[t]{0.32\textwidth}
%   \centering
%   \captionsetup{justification=centering}
%     \includegraphics[scale=0.165]{fig/wiki_c3_tk.png}
%   \caption{$C_3$ Target Language Group}
%   \label{fig:wiki_c3}
% \end{subfigure}
%     \hfill
% \begin{subfigure}[t]{0.32\textwidth}
%   \centering
%   \captionsetup{justification=centering}
%     \includegraphics[scale=0.165]{fig/wiki_th_tk.png}
%   \caption{Thai ({\small TH})}
%   \label{fig:wiki_th}
% \end{subfigure}
% \caption{Delta between the few-shot results of reported methods and {\small RAND} on NER task in terms of the total number of tokens selected for each $k$ with mBERT. The delta is calculated by projecting the few-shot results of {\small RAND}. We observe the similar trends in $C_1$ and $C_2$, so we are only showing the results for $C_1$. We observe that methods based on PU performs better than to that of PE. The X-axis has been shown in log-scale.} 
% \label{fig:ner_result}
% \end{figure*}

\subsection{Few-Shot Transfer}
We add $k$ additional examples from a target language and report the improvement of few-shot performance over the zero-shot performance reported in Section 3.3, where $k$ examples are chosen according to the proposed strategies in Section 3, namely random sampling ({\small RAND}), DCE, PE, GE, and LE. We use similar training and evaluation setups for the few-shot transfer experiments as we used in the zero-shot setting and repeat the experiments with three random seeds. We consider three {\small RAND} baselines and report the average for all the data selection experiments.

\section{Results}
We calculated the difference between the $\text{F}_1$-scores of few-shot and zero-shot setups, {\em delta}s($\triangle$), for each language separately, but we observed different sampling strategies to work better depending upon the cross-lingual transfer gap. Therefore, we present the experimental results after categorizing languages by the transfer gap as indicated by the zero-shot performance, shown in Table \ref{tbl:zero_shot}. We categorize the languages in three groups: $C_1$, $C_2$ and $C_3$, and are coloured as light grey, dark grey and very dark grey respectively in Table \ref{tbl:zero_shot}. For NER task, groups are defined as $C_1$ $\in$ \{{\small BG, DE, EL, ES, EU, FI, FR, HI, RU, SV, TR, VI}\}, $C_2$ $\in$ \{{\small AR, HE, JA, KO, UR, ZH}\}, and $C_3$ $\in$ \{{\small TH}\}. For POS, groups are defined as $C_1$ $\in$ \{{\small BG, DE, ES, EL, FI, RU, SV, TR}\}, and $C_2$ $\in$ \{{\small AR, EU, HE, HI, JA, KO, JA, UR, ZH}\}. For XNLI, the groups are different for XLM-R and mBERT, hence we have mentioned them in the Appendix.

We report the {\em delta}s, for NER and POS tasks in Table \ref{tbl:few_shot_ner_en} and \ref{tbl:few_shot_pos_en}, respectively. The reported deltas are averaged across all the target languages for each language group. All the reported values are positive, which means in all cases,  performance for the few-shot is higher than that for the zero-shot.
The proposed methods require two parameters $\lambda$ and $\gamma$ for data selection. We perform experiments for $\lambda \in \{0, 0.5, 1\}$ and $\gamma \in \{1, 2, 3\}$, which are reported in Appendix \ref{sec:ablation}. In Table~\ref{tbl:few_shot_ner_en} and \ref{tbl:few_shot_pos_en}, we are reporting the best setups for PE and LE, where we observe the highest gains for NER and POS tasks, respectively.
% We also show the few-shot performance in terms of the total number of tokens selected for each $k$ in Figure $\ref{fig:token_result}$.
The methods based on PE and LE consistently outperform the baseline {\small RAND} and DCE for all values of $k$ on POS task, and most of the cases in the case of NER. %providing average absolute gains of 1.5, 1.4 and 0.8 points for categories $C_1$, $C_2$ and $C_3$, respectively with values of $k$ less than 1000. 
%On POS task, both PE and PU provide average absolute gains of 0.8, 0.9, and 1.3 points for each category with values of $k$ less than 1000. 
DCE performs worse compared to {\small RAND} for all the languages from groups $C_1$ and $C_2$. In general, the gains obtained through PE and LE compared to {\small RAND} are higher for $C_1$ than $C_2$. Similarly, the proposed approaches are more useful compared to {\small RAND} for small values $k$, and the advantage of our sampling strategies diminishes as $k$ approaches to 1000. For {\small TH} ($ \in C_3$), due to deficient zero-shot transfer performance in NER, the gains are not consistent across models. However, all three approaches outperform {\small RAND} for small values of $k$.

% a different combination of $\lambda$ and $\gamma$ gives the highest gains, and hence, it is not included in the results reported in Table \ref{tbl:few_shot_ner_en}. In Figure \ref{fig:wiki_th}, we can see that DCE provides the best gains for {\small TH} compared to other methods in terms of the number of tokens. The gains obtained through PU are also comparable to DCE. 

% \begin{figure}[h!]
%     \centering
%     \includegraphics[scale=0.19]{fig/wiki_th_tk.png} 
% % \begin{figure}[h!]
% %     \centering
% %     \begin{subfigure}[t]{0.23\textwidth}
% %         \centering
% %         \includegraphics[scale=0.135]{fig/wiki_th_tk.png}
% %          \caption{TH (mBERT)}
% %         \label{fig:wiki_th_mbert}
% %     \end{subfigure}
% %         \hfill
% %     \begin{subfigure}[t]{0.23\textwidth}
% %         \centering
% %         \includegraphics[scale=0.135]{fig/wiki_th_tk_xlmr.png}
% %         \caption{TH (XLM-R)}
% %         \label{fig:wiki_th_xlmr}
% %     \end{subfigure}
% %     % \vspace{-0.5em}
%   \caption{Few-shot results on the Thai language for NER task in terms of the total number of tokens selected for each k.
% %   The data selection based on DCE and PU is superior to random sampling in terms of the number of tokens selected.
%   }
%     \label{fig:wiki_th}
%     % \vspace{-1em}
% \end{figure}

\begin{table}[h]
\centering
\small
\setlength\tabcolsep{5.5pt}%
\begin{tabular}{c|c|cccccc}
\toprule
\multicolumn{1}{c}{}  & \textbf{Method} & $10$ & $100$ & $500$ & $1k$ & $5k$ & $10k$\\ 
\midrule
 & RAND & -0.5 &
-0.9 &
-0.2 &
0.6 &
2.4 &
3.4
 \\ 
 & DCE & 0.0 &
-0.1 &
-0.1 &
0.5 &
2.8 &
3.9
 \\
  & PE ($\lambda=1$) & 
  -0.4 &
-1.0 &
-0.3 &
0.4 &
2.6 &
3.4
\\
 & GE & -0.1 &
-0.9 &
-0.5 &
0.2 &
2.6 &
3.1
\\
\multirow{-5}{*}{\rotatebox[origin=c]{90}{mBERT}} & GE ($\gamma =1$) & 
\textbf{0.1} &
\textbf{-0.1} &
\textbf{0.7} &
\textbf{1.4} &
\textbf{3.3} &
\textbf{4.2}
 \\ 
\midrule
 & RAND & -0.1 &
-0.2 &
0.0 &
0.2 &
1.2 &
1.5
 \\ 
 & DCE & \textbf{0.2} &
\textbf{0.7} &
0.5 &
0.7 &
1.6 &
{2.1}
 \\
   & PE ($\lambda=1$) & 
   -0.4 &
-0.0 &
0.0 &
0.3 &
1.3 &
1.5
\\
  & GE & -0.3 &
-0.1 &
0.2 &
0.4 &
1.5 &
\textbf{2.6}
\\
\multirow{-4}{*}{\rotatebox[origin=c]{90}{XLM-R}} & GE ($\gamma =1$) &
0.1 &
0.5 &
\textbf{0.6} &
\textbf{1.2} &
\textbf{1.9} &
{2.1}
\\
\bottomrule
\end{tabular} 
% \vspace{-0.5em}
\caption {Averaged few-shot performance on XNLI tasks with varying number of target language examples $k$ using {\small EN} as the pivot language.}
\label{tbl:few_shot_xnli_en}
% \vspace{-0.5em}
\end{table}

For XNLI, the averaged deltas across all languages are reported in Table \ref{tbl:few_shot_xnli_en}. As DCE requires a sentence to train n-gram language model, hence we represent a sentence in XNLI by joining the hypothesis and the premise of an instance with a separator (-). The few-shot improvements are less pronounced than the sequence labeling tasks; noticeable gains start after seeing $k=500$ target-language examples. As the size of the target-language corpus in XNLI is enormous compared to POS and NER, we also evaluated the methods for $k=10000$. Surprisingly, GE ($\gamma=1$) and DCE outperforms {\small RAND}. As DCE selects examples in batches of 10, it selects examples having similar contexts similar to GE ($\gamma=1$), which benefits the few-shot learning. Since GE ($\gamma=1$) also includes uncertainty sampling, it outperforms DCE for most of the values of $k$.  Due to the large corpus size of XNLI, diversity becomes crucial during sampling. We observe low few-shot gains for PE as it does not induce diversity. To measure the impact of pivot size, we trained a zero-shot model with 40k {\small EN} examples and observe similar trends for both DCE and GE (see Table \ref{tbl:few_shot_xnli_en_full} in Appendix).

\begin{table}[ht]
\centering
\small
\setlength\tabcolsep{6pt}%
\begin{tabular}{c|cc|cc|cc}
\toprule
 & \multicolumn{2}{c}{XNLI}  & \multicolumn{2}{|c}{NER} & \multicolumn{2}{|c}{POS} 
\\
\cmidrule(lr){2-3}\cmidrule(lr){4-5}\cmidrule(lr){6-7} 
 \multirow{-1}{*}{\textbf{Method}} & \thead{${C_1}$ \\ (24)}  & \thead{${C_2}$ \\ (4)} & \thead{${C_1}$ \\(24) } & \thead{${C_2}$ \\(12)} & \thead{${C_1}$ \\ (16)} & \thead{${C_2}$ \\ (16)  }
\\ 
\midrule
PE & 1 & 1 & 8 & 1 & 1 & 7
\\
LE & - & -& \textbf{15} &  \textbf{5} & \textbf{3} & \textbf{10}
\\
GE & \textbf{18} & \textbf{2} & - & - & - & -
\\
\bottomrule
\end{tabular}
% \vspace{-0.5em}
\caption {Pairwise $t$-Test is performed using the proposed methods against {\small RAND}. We have reported the number of languages in each group having significance level of 0.1 using both XLM-R and mBERT models.}
\label{tbl:ttest}
\vspace{-0.5em}
\end{table}

\subsection{Effect of $\lambda$ and $\gamma$ parameters}
In Appendix B, we have provided detailed results by varying $\lambda$ and $\gamma$. For LE, a higher value of $\lambda$ is required for the POS task due to the higher number of class labels than NER. The number of classes is 18 for POS and 7 for the NER task. Due to the higher number of class labels, the norm of loss embedding distribution has a higher tail. Hence, a higher value of $\lambda$ is required for POS. We limited the value of $\lambda$ to $0.5$ as beyond that, very few examples were left for selection.

We incorporate $\gamma$ parameter to include examples similar in context. As sentences with similar context will also have similar class labels in the case of POS and NER tasks, further decreasing the diversity in samples. Hence, we only consider experimenting with $\gamma$ for XNLI. We observe that $\gamma=1$ provides the best performance on average, suggesting that having two samples of similar contexts provides better few-shot learning. 
% For values greater than 1, gains start diminishing due to lower diversity.

\subsection{Statistical Significance Test}
We perform a pairwise $t$-Test for measuring the statistical significance of the proposed methods against the {\small RAND} baseline. We perform $t$-test for each language using both mBERT and XLM-R and have reported the number of languages having $p$-value less than the critical point (which is 0.1 in our case) for each language group. We have considered the following methods in our tests: GE ($\gamma = 1$) for XNLI, LE for NER, LE ($\lambda = 0.5$) for POS and PE ($\lambda=1$) for all the tasks.

In Table \ref{tbl:ttest}, we notice that for XNLI, GE provides significant gains than {\small RAND} for 18 out of 24 cases from $C_1$ group, and 2 out of 4 cases from $C_2$ group. For NER, the gains are significant for 20 out of 32 cases while using LE, but only 9 cases have significant gains using PE. For POS, we observe LE provide significant gains for cases compared to PE. We can conclude that the embedding-based methods provide better gains than uncertainty-based methods for most languages.

% 15 cases out of 24 to be significant from $C_1$ group, while only 5 out of 12 cases from $C_2$ group and 1 out of 2 from $C_3$ group. This is due to the saturation of few-shot gains for higher values of $k$ in NER. 
% For POS, only 3 out of 16 cases are found to be significant from $C_1$ group, however 10 out of 16 cases from $C_2$ group have significant few-shot gains. 

\begin{table}[h]
\centering
\small
\setlength\tabcolsep{5.5pt}%
\begin{tabular}{c|c|ccccc}
\toprule
\multicolumn{1}{c}{}  & \textbf{Method} & $10$ & $50$ & $100$ & $500$ & $1000$\\ 
\midrule
 & RAND & 3.9 &
8.8 &
10.5 &
18.6 &
21.2
 \\ 
 & DCE & 1.8 &
5.5 &
6.5 &
15.3 &
19.2
\\
\multirow{-3}{*}{\rotatebox[origin=c]{90}{mBERT}} & LE & 
\textbf{5.0} &
\textbf{11.0} &
\textbf{12.2} &
\textbf{19.8} &
\textbf{22.0}
 \\ 
\midrule
 & RAND & 7.3 &
15.3 &
\textbf{17.1} &
\textbf{26.5} &
\textbf{29.3}
 \\ 
 & DCE & -0.8 &
8.6 &
10.5 &
22.9 &
26.9
 \\
\multirow{-3}{*}{\rotatebox[origin=c]{90}{XLM-R}} & LE &
\textbf{8.7} &
\textbf{15.9} &
\textbf{17.1} &
{26.1} &
29.1
\\
\bottomrule
\end{tabular} 
% \vspace{-0.5em}
\caption {$\triangle$ Delta between Few-shot and zero-shot performance on NER tasks using \textbf{{\small ZH}} as the pivot language, averaged across all languages.}
\label{tbl:few_shot_ner_zh}
\vspace{-0.5em}
\end{table}

\section{Impact of Pivot Language}
\label{sec:pivot_data}
We conduct few-shot experiments considering {\small ZH} as the pivot language to validate the effectiveness of our method across different pivots. The delta between the gains using {\small ZH} as the pivot have been reported in Table \ref{tbl:few_shot_ner_zh} on the NER task. The delta has been averaged across all the languages. LE provides consistent gains over {\small RAND}, and gains saturate beyond 500 examples.
 
\subsection{Embedding Visualization}
\begin{figure}[!ht]
    \centering
    \includegraphics[scale=0.35]{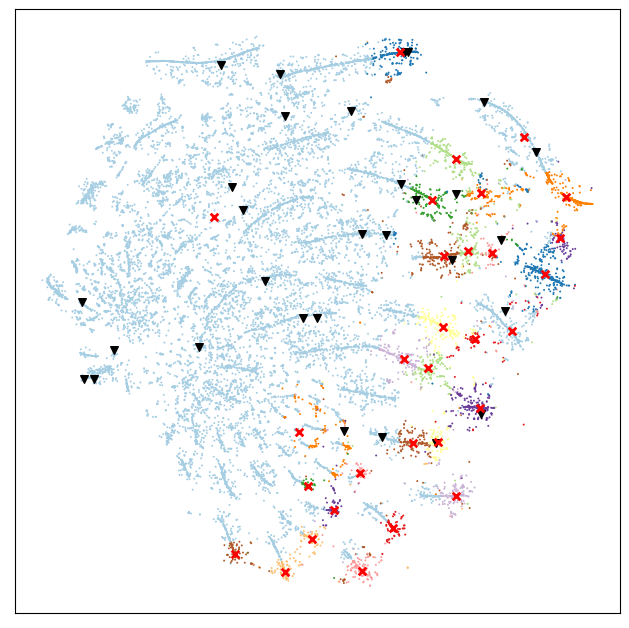}
\vspace{-0.5em}
  \caption{The t-SNE visualization of loss embedding using the complete {\small DE} language NER corpus with XLM-R. The clusters of loss embedding are highlighted with {\color{red} $\times$}, while samples from {\small RAND} are highlighted with {\color{black} $\blacktriangledown$}.
%   We notice that examples having lower norm of loss embedding are clustered together and highlighted with ocean colour. This shows that the samples via LE method have higher diversity and uncertainty.
  }
    \label{fig:ner_de_vis}
    \vspace{-0.5em}
\end{figure}

We visualize the loss embeddings for {\small DE} language using  t-SNE \cite{van2008visualizing} in Figure \ref{fig:ner_de_vis}. Most of the samples using {\small RAND} ({\color{black} $\blacktriangledown$}) tend to have lower norm of loss embedding, which may not be ideal for few-shot learning. We notice that examples having lower norm of loss embedding are clustered together and highlighted with ocean colour. Hence, samples selected via LE ({\color{red} $\times$}) are more likely to have higher norm or higher uncertainty estimates. It is also evident that the samples from LE (cluster centre) will have higher diversity than {\small RAND} for few-shot learning.

\begin{table*}[ht]
\centering
\scriptsize
\setlength\tabcolsep{1.5pt}%
\begin{tabular}{c|P{0.48\linewidth}|P{0.48\linewidth}}
\toprule
 &  Raw Text & Translated Text\\ 
\midrule
 & Er beschäftigt sich dort hauptsächlich mit dem \hl{Auswärtigen} \hl{Amt} und der \hl{SPD}. & There he mainly deals with the \hl{Foreign} \hl{Office} and the \hl{SPD}.
\\
 & In der weiblichen Hauptrolle ist \hl{Elżbieta} \hl{Czyżewska} zu sehen. & \hl{Elżbieta} \hl{Czyżewska} stars in the female lead.
\\
 & Aus \hl{Asien} in den Nordwesten , wo die Erfindung lange verharrte , dann an die Ostküste gelangte , um erst rund drei Jahrtausende später den Westen zu erreichen. & From \hl{Asia} to the northwest, where the invention remained for a long time, then reached the east coast, only to reach the west some three millennia later.
 \\
\multirow{-7}{*}{\rotatebox[origin=c]{90}{{\small RAND}}} & Er lebt in \hl{Fernwald}. & He lives in \hl{Fernwald}.
\\
\midrule
 & Auch unter Trainer \hl{Joachim} \hl{Löw} behielt er den Posten des Managers. & He also retained the post of manager under coach \hl{Joachim} \hl{Löw}.
\\
 & \hl{Maria} \hl{de} \hl{Lourdes} \hl{Ruivo} \hl{da} \hl{Silva} \hl{Matos} \hl{Pintasilgo}. & \hl{Maria} \hl{de} \hl{Lourdes} \hl{Ruivo} \hl{da} \hl{Silva} \hl{Matos} \hl{Pintasilgo}.
\\
 & Weiterleitung \hl{Atlantic} \hl{Coast} \hl{Hockey} \hl{League}. & Forwarding \hl{Atlantic} \hl{Coast} \hl{Hockey} \hl{League}.
\\
\multirow{-4}{*}{\rotatebox[origin=c]{90}{LE}} & \hl{Chuck} \hl{Weyant} und \hl{Al} \hl{Herman} gingen mit Dunn-Rennwagen an den Start , wobei der 13. & \hl{Chuck} \hl{Weyant} and Al \hl{Herman} competed in Dunn racers, with 13th.
 \\
\bottomrule
\end{tabular}
\vspace{-0.5em}
\caption {Samples in {\small DE} language from NER task using {\small RAND} and LE methods for $k=10$ using XLM-R. Highlighted tokens are entities. We observe that LE tends to pick examples containing more entities than {\small RAND}.}
\label{tbl:examples}
\vspace{-0.5em}
\end{table*}

 \begin{table}[ht]
\centering
\small
\setlength\tabcolsep{4pt}%
\begin{tabular}{c|cc}
\toprule
Language &  {\small RAND} & LE \\ 
\midrule
ar & 30.15\% & \textbf{35.21\%} 
\\
eu & 16.53\% & \textbf{20.17\%}
\\
he & 34.11\% & \textbf{34.77\%} 
\\
hi & 28.25\% & \textbf{31.70\%} 
\\
ko & 31.65\% & \textbf{34.64\%} 
\\
ja & 86.34\% &  \textbf{86.44\%} 
\\
ur & \textbf{20.29\%} & 20.26\% 
\\
zh & 73.27 & \textbf{77.52} 
\\
\bottomrule
\end{tabular}
\caption {Percentage of tokens from sentences sampled using {\small RAND} and LE($\lambda=0.5$) for $k=10$ from POS task. We calculated the percentage of tokens from class labels that are mispredicted using the zero-shot model for more than 40\% on the whole language corpus. We observe that the LE method selects sentences containing more incorrect class labels without access to the ground truth labels. The languages from $C_1$ group are not considered as the gains are not relatively low.}
\label{tbl:misrate_pos}
% \vspace{-0.5em}
\end{table}

\subsection{Qualitative Analysis of Samples}
We have compared sentences selected using {\small RAND} and LE for the NER task in Table \ref{tbl:examples}. Random sampling has no constraints due to which it may select examples having very few entities which might not improve the few-shot performance. Since LE uses loss as the measure of uncertainty, it selects sentences with a higher number of entities often miss-classified by the zero-shot model. Hence, LE will most probably improve the few-shot performance compared to {\small RAND}. Similarly, we observe in Table \ref{tbl:misrate_pos} that for POS, LE selects sentences containing class labels that are often incorrectly tagged by the zero-shot model. In the case of XNLI, we found that the GE-based method selects more competing examples (similar hypotheses for different premises leading to different labels), which effectively can enhance the model capability. Selected examples using different methods for POS and XNLI tasks can be found in Appendix \ref{sec:qualitative}.

\section{Related Work}
\subsection{Cross-lingual Transfer}
In recent years, several pre-trained multilingual language models have been proposed including mBERT \cite{devlin-etal-2019-bert}, XLM \cite{NEURIPS2019_c04c19c2}, XLM-R \cite{conneau2020unsupervised}, mBART~\cite{liu2020multilingual}, mT5~\cite{xue2020mt5} and ERNIE-M~\cite{ouyang2020ernie} for cross-lingual transfer. \citet{pires-etal-2019-multilingual} show mBERT to have good zero-shot performance on NER and POS tagging tasks and attributed the effectiveness of transfer to the typological similarity between the languages. In contrast, several works \cite{karthikeyan2019cross, wu-dredze-2019-beto}
% artetxe-etal-2020-cross}
have shown that cross-lingual transfer does not depend on subword vocabulary overlap and joint training across languages.~\citet{lauscher2020zero} empirically demonstrate that both pre-training corpora sizes and linguistic similarity are strongly correlated with the zero-shot transfer. Target languages with smaller pretraining corpora or higher linguistic dissimilarity with the pivot language have a low zero-shot transfer. Furthermore, they have shown that the gap can be reduced significantly by fine-tuning with a small number of target-language examples. \citet{nooralahzadeh-etal-2020-zero} study the cross-lingual transfer in meta-learning setting and demonstrate improvement in zero-shot and few-shot settings. While \cite{lauscher2020zero, nooralahzadeh-etal-2020-zero} focus on reducing zero-shot transfer gap using few-shot learning, in this work, we explore the data selection methods to get better cross-lingual transfer than the often used random sampling.

\subsection{Training Data Selection}
The problem of training data selection has been extensively studied for several NLP tasks, with the most notable ones from area of Machine Translation systems where target-domain data is limited and large non-domain-specific data is available. The task is to pick sentences that are closer to the target domain and also penalize the sentences which are out-of-domain. \citet{moore-lewis-2010-intelligent} and \citet{axelrod-etal-2011-domain} address this problem by ranking sentences using the cross-entropy of target-domain-specific and non-domain-specific $n$-gram language models. \citet{ dara2014active} employ an extension of the cross-entropy difference by including a vocabulary saturation filter which removes selection of very similar sentences. \citet{song-etal-2012-entropy} have shown the effectiveness of cross-entropy selecting in-domain data for word segmentation and POS tagging tasks. We also use an extension of cross-entropy for selecting training data from the target language corpus for effective few-shot transfer using multilingual transformer models and compare with the proposed methods. 

\subsection{Active Learning}
Active Learning has been widely used to reduce the amount of labeling to learn good models, \cite{yoo2019learning, fu2013survey}. Uncertainty sampling methods have been commonly used in AL, where the most uncertain samples are selected for labeling. Various metrics have defined uncertainty using least confidence, sample margin, and predictive entropy. On the other hand, diversity sampling methods \cite{sener2018active, gissin2019discriminative} select examples which can act as a surrogate for the entire dataset. 
% Recently, \citet{ash2019deep} proposed {\small BADGE} which combines both uncertainty and diversity sampling.
% \cite{holub2008entropy, zhang2019ensemble} have used the entropy of the class probabilities to select examples.
% Recently \citet{gal2017deep} have used uncertainty estimates provided by MC-Dropout \cite{gal2016dropout} to improve sample selection in AL. 
\citet{ chaudhary2019little} used AL-based approaches to select entity spans for labeling in a cross-lingual transfer learning setting. However, this work was limited to only two languages. Our work focuses on data selection for cross-lingual transfer on a large and diverse set of target languages.

\section{Discussion and Conclusion}
This work explored various data sampling strategies for few-shot learning for two sequence labeling and a semantic tasks on 20 target languages. Our study shows that the embedding-based strategies, LE and GE, consistently outperform random sampling baseline across languages and sample sizes. Some of the salient observations are as follows. On NER and POS tasks, languages of the group $C_2$ show significant improvements in few-shot performance, suggesting that the gains from few-shot learning are strongly correlated to the zero-shot transfer gap. LE and GE-based data selection methods show consistent gains over the {\small RAND} strategy for each target language group, but these gains saturate as the sample size, $k$, increases beyond 500. The saturation occurs due to the relatively smaller target-language corpus size (varies between 5k and 20k for NER and POS, respectively) effectively reducing the diversity in the total sample. LE provides better few-shot performance than PE in terms of statistical significance. DCE only performs better than {\small RAND} for Thai. As DCE does not use any form of information from the fine-tuned model and if the target-language corpus size is small, it fails to select novel target language examples any better than {\small RAND}. However, in {\small TH}, for which zero-shot performance is close to zero, DCE selects the highly representative and diverse training examples for small values of $k$. The trends for XNLI are different from that of the sequence labeling tasks. GE and DCE outperform all other methods, with gains increasing with the value of $k$, which suggests that the size of the target-language corpus is crucial for data selection. XNLI has about 400k examples in each target-language corpus, much larger than that of NER and POS, signifying the importance of diversity sampling.

Based on our observations, we recommend the LE-based sampling strategy for data selection for cross-lingual few-shot transfer for sequence labeling tasks and GE-based sampling for classification tasks. While the optimal parameter setting for the LE sampling algorithm varies across tasks, we recommend the vanilla LE method without any parameter for most of the tasks. For tasks having higher number of class labels, we recommend using LE variant with $\lambda$ such as 0 or 0.5.

% we do observe a few patterns which can be distilled into the following two thumb-rules: (a) $\gamma = 0.5$ works for most situations; (b) $\lambda$ should be set at 1, 0.5 and 0 respectively for high (i.e., when transfer gap is less than 30\%), moderate (transfer gap between 30\% and 50\%) and low (transfer gap less more than 50\%) zero-shot transfer performance in the target language. 

In future  the work can be extended to other high-level tasks such as cross-lingual QA and Machine translation. We would also like to extend this work in a reinforcement learning \cite{liu2019reinforced} or meta-learning \cite{tseng2020cross} framework, where the parameters can be automatically learnt for various tasks and settings.

\bibliographystyle{acl_natbib}
\bibliography{custom}

\begin{thebibliography}{38}
\expandafter\ifx\csname natexlab\endcsname\relax\def\natexlab#1{#1}\fi

\bibitem[{Artetxe et~al.(2020)Artetxe, Ruder, and
  Yogatama}]{artetxe-etal-2020-cross}
Mikel Artetxe, Sebastian Ruder, and Dani Yogatama. 2020.
\newblock \href {https://doi.org/10.18653/v1/2020.acl-main.421} {On the
  cross-lingual transferability of monolingual representations}.
\newblock In \emph{Proceedings of the 58th Annual Meeting of the Association
  for Computational Linguistics}, pages 4623--4637, Online. Association for
  Computational Linguistics.

\bibitem[{Arthur and Vassilvitskii(2006)}]{arthur2006k}
David Arthur and Sergei Vassilvitskii. 2006.
\newblock k-means++: The advantages of careful seeding.
\newblock Technical report, Stanford.

\bibitem[{Ash et~al.(2019)Ash, Zhang, Krishnamurthy, Langford, and
  Agarwal}]{ash2019deep}
Jordan~T Ash, Chicheng Zhang, Akshay Krishnamurthy, John Langford, and Alekh
  Agarwal. 2019.
\newblock Deep batch active learning by diverse, uncertain gradient lower
  bounds.
\newblock In \emph{International Conference on Learning Representations}.

\bibitem[{Axelrod et~al.(2011)Axelrod, He, and Gao}]{axelrod-etal-2011-domain}
Amittai Axelrod, Xiaodong He, and Jianfeng Gao. 2011.
\newblock \href {https://www.aclweb.org/anthology/D11-1033} {Domain adaptation
  via pseudo in-domain data selection}.
\newblock In \emph{Proceedings of the 2011 Conference on Empirical Methods in
  Natural Language Processing}, pages 355--362, Edinburgh, Scotland, UK.
  Association for Computational Linguistics.

\bibitem[{Blitzer et~al.(2007)Blitzer, Dredze, and
  Pereira}]{blitzer2007biographies}
John Blitzer, Mark Dredze, and Fernando Pereira. 2007.
\newblock Biographies, bollywood, boom-boxes and blenders: Domain adaptation
  for sentiment classification.
\newblock In \emph{Proceedings of the 45th annual meeting of the association of
  computational linguistics}, pages 440--447.

\bibitem[{Chaudhary et~al.(2019)Chaudhary, Xie, Sheikh, Neubig, and
  Carbonell}]{chaudhary2019little}
Aditi Chaudhary, Jiateng Xie, Zaid Sheikh, Graham Neubig, and Jaime~G
  Carbonell. 2019.
\newblock A little annotation does a lot of good: A study in bootstrapping
  low-resource named entity recognizers.
\newblock In \emph{Proceedings of the 2019 Conference on Empirical Methods in
  Natural Language Processing and the 9th International Joint Conference on
  Natural Language Processing (EMNLP-IJCNLP)}, pages 5164--5174.

\bibitem[{Conneau et~al.(2020)Conneau, Khandelwal, Goyal, Chaudhary, Wenzek,
  Guzm{\'a}n, Grave, Ott, Zettlemoyer, and Stoyanov}]{conneau2020unsupervised}
Alexis Conneau, Kartikay Khandelwal, Naman Goyal, Vishrav Chaudhary, Guillaume
  Wenzek, Francisco Guzm{\'a}n, {\'E}douard Grave, Myle Ott, Luke Zettlemoyer,
  and Veselin Stoyanov. 2020.
\newblock Unsupervised cross-lingual representation learning at scale.
\newblock In \emph{Proceedings of the 58th Annual Meeting of the Association
  for Computational Linguistics}, pages 8440--8451.

\bibitem[{Conneau and Lample(2019)}]{NEURIPS2019_c04c19c2}
Alexis Conneau and Guillaume Lample. 2019.
\newblock \href
  {https://proceedings.neurips.cc/paper/2019/file/c04c19c2c2474dbf5f7ac4372c5b9af1-Paper.pdf}
  {Cross-lingual language model pretraining}.
\newblock In \emph{Advances in Neural Information Processing Systems},
  volume~32. Curran Associates, Inc.

\bibitem[{Conneau et~al.(2018)Conneau, Rinott, Lample, Williams, Bowman,
  Schwenk, and Stoyanov}]{conneau2018xnli}
Alexis Conneau, Ruty Rinott, Guillaume Lample, Adina Williams, Samuel Bowman,
  Holger Schwenk, and Veselin Stoyanov. 2018.
\newblock Xnli: Evaluating cross-lingual sentence representations.
\newblock In \emph{Proceedings of the 2018 Conference on Empirical Methods in
  Natural Language Processing}, pages 2475--2485.

\bibitem[{Dandapat et~al.(2009)Dandapat, Biswas, Choudhury, and
  Bali}]{dandapat-etal-2009-complex}
Sandipan Dandapat, Priyanka Biswas, Monojit Choudhury, and Kalika Bali. 2009.
\newblock \href {https://www.aclweb.org/anthology/W09-3002} {Complex linguistic
  annotation {--} no easy way out! a case from {B}angla and {H}indi {POS}
  labeling tasks}.
\newblock In \emph{Proceedings of the Third Linguistic Annotation Workshop
  ({LAW} {III})}, pages 10--18, Suntec, Singapore. Association for
  Computational Linguistics.

\bibitem[{Dara et~al.(2014)Dara, van Genabith, Liu, Judge, and
  Toral}]{dara2014active}
Aswarth~Abhilash Dara, Josef van Genabith, Qun Liu, John Judge, and Antonio
  Toral. 2014.
\newblock Active learning for post-editing based incrementally retrained mt.
\newblock In \emph{Proceedings of the 14th Conference of the European Chapter
  of the Association for Computational Linguistics, volume 2: Short Papers},
  pages 185--189.

\bibitem[{Devlin et~al.(2019)Devlin, Chang, Lee, and
  Toutanova}]{devlin-etal-2019-bert}
Jacob Devlin, Ming-Wei Chang, Kenton Lee, and Kristina Toutanova. 2019.
\newblock \href {https://doi.org/10.18653/v1/N19-1423} {{BERT}: Pre-training of
  deep bidirectional transformers for language understanding}.
\newblock In \emph{Proceedings of the 2019 Conference of the North {A}merican
  Chapter of the Association for Computational Linguistics: Human Language
  Technologies, Volume 1 (Long and Short Papers)}, pages 4171--4186,
  Minneapolis, Minnesota. Association for Computational Linguistics.

\bibitem[{Fort(2016)}]{fort2016collaborative}
Karën Fort. 2016.
\newblock Collaborative annotation for reliable natural language processing:
  Technical and sociological aspects.

\bibitem[{Fu et~al.(2013)Fu, Zhu, and Li}]{fu2013survey}
Yifan Fu, Xingquan Zhu, and Bin Li. 2013.
\newblock A survey on instance selection for active learning.
\newblock \emph{Knowledge and information systems}, 35(2):249--283.

\bibitem[{Gissin and Shalev-Shwartz(2019)}]{gissin2019discriminative}
Daniel Gissin and Shai Shalev-Shwartz. 2019.
\newblock Discriminative active learning.
\newblock \emph{arXiv preprint arXiv:1907.06347}.

\bibitem[{Joshi et~al.(2020)Joshi, Santy, Budhiraja, Bali, and
  Choudhury}]{joshi-etal-2020-state}
Pratik Joshi, Sebastin Santy, Amar Budhiraja, Kalika Bali, and Monojit
  Choudhury. 2020.
\newblock \href {https://doi.org/10.18653/v1/2020.acl-main.560} {The state and
  fate of linguistic diversity and inclusion in the {NLP} world}.
\newblock In \emph{Proceedings of the 58th Annual Meeting of the Association
  for Computational Linguistics}, pages 6282--6293, Online. Association for
  Computational Linguistics.

\bibitem[{Karthikeyan et~al.(2019)Karthikeyan, Wang, Mayhew, and
  Roth}]{karthikeyan2019cross}
K~Karthikeyan, Zihan Wang, Stephen Mayhew, and Dan Roth. 2019.
\newblock Cross-lingual ability of multilingual bert: An empirical study.
\newblock In \emph{International Conference on Learning Representations}.

\bibitem[{Lauscher et~al.(2020)Lauscher, Ravishankar, Vuli{\'c}, and
  Glava{\v{s}}}]{lauscher2020zero}
Anne Lauscher, Vinit Ravishankar, Ivan Vuli{\'c}, and Goran Glava{\v{s}}. 2020.
\newblock From zero to hero: On the limitations of zero-shot language transfer
  with multilingual transformers.
\newblock In \emph{Proceedings of the 2020 Conference on Empirical Methods in
  Natural Language Processing (EMNLP)}, pages 4483--4499.

\bibitem[{Liu et~al.(2019)Liu, Song, Zou, and Zhang}]{liu2019reinforced}
Miaofeng Liu, Yan Song, Hongbin Zou, and Tong Zhang. 2019.
\newblock Reinforced training data selection for domain adaptation.
\newblock In \emph{Proceedings of the 57th annual meeting of the association
  for computational linguistics}, pages 1957--1968.

\bibitem[{Liu et~al.(2020)Liu, Gu, Goyal, Li, Edunov, Ghazvininejad, Lewis, and
  Zettlemoyer}]{liu2020multilingual}
Yinhan Liu, Jiatao Gu, Naman Goyal, Xian Li, Sergey Edunov, Marjan
  Ghazvininejad, Mike Lewis, and Luke Zettlemoyer. 2020.
\newblock Multilingual denoising pre-training for neural machine translation.
\newblock \emph{Transactions of the Association for Computational Linguistics},
  8:726--742.

\bibitem[{Moore and Lewis(2010)}]{moore-lewis-2010-intelligent}
Robert~C. Moore and William Lewis. 2010.
\newblock \href {https://www.aclweb.org/anthology/P10-2041} {Intelligent
  selection of language model training data}.
\newblock In \emph{Proceedings of the {ACL} 2010 Conference Short Papers},
  pages 220--224, Uppsala, Sweden. Association for Computational Linguistics.

\bibitem[{Nivre et~al.(2016)Nivre, de~Marneffe, Ginter, Goldberg, Haji{\v{c}},
  Manning, McDonald, Petrov, Pyysalo, Silveira, Tsarfaty, and
  Zeman}]{nivre-etal-2016-universal}
Joakim Nivre, Marie-Catherine de~Marneffe, Filip Ginter, Yoav Goldberg, Jan
  Haji{\v{c}}, Christopher~D. Manning, Ryan McDonald, Slav Petrov, Sampo
  Pyysalo, Natalia Silveira, Reut Tsarfaty, and Daniel Zeman. 2016.
\newblock \href {https://www.aclweb.org/anthology/L16-1262} {{U}niversal
  {D}ependencies v1: A multilingual treebank collection}.
\newblock In \emph{Proceedings of the Tenth International Conference on
  Language Resources and Evaluation ({LREC}'16)}, pages 1659--1666,
  Portoro{\v{z}}, Slovenia. European Language Resources Association (ELRA).

\bibitem[{Nooralahzadeh et~al.(2020)Nooralahzadeh, Bekoulis, Bjerva, and
  Augenstein}]{nooralahzadeh-etal-2020-zero}
Farhad Nooralahzadeh, Giannis Bekoulis, Johannes Bjerva, and Isabelle
  Augenstein. 2020.
\newblock \href {https://doi.org/10.18653/v1/2020.emnlp-main.368} {Zero-shot
  cross-lingual transfer with meta learning}.
\newblock In \emph{Proceedings of the 2020 Conference on Empirical Methods in
  Natural Language Processing (EMNLP)}, pages 4547--4562, Online. Association
  for Computational Linguistics.

\bibitem[{Ouyang et~al.(2020)Ouyang, Wang, Pang, Sun, Tian, Wu, and
  Wang}]{ouyang2020ernie}
Xuan Ouyang, Shuohuan Wang, Chao Pang, Yu~Sun, Hao Tian, Hua Wu, and Haifeng
  Wang. 2020.
\newblock Ernie-m: Enhanced multilingual representation by aligning
  cross-lingual semantics with monolingual corpora.
\newblock \emph{arXiv preprint arXiv:2012.15674}.

\bibitem[{Pires et~al.(2019)Pires, Schlinger, and
  Garrette}]{pires-etal-2019-multilingual}
Telmo Pires, Eva Schlinger, and Dan Garrette. 2019.
\newblock \href {https://doi.org/10.18653/v1/P19-1493} {How multilingual is
  multilingual {BERT}?}
\newblock In \emph{Proceedings of the 57th Annual Meeting of the Association
  for Computational Linguistics}, pages 4996--5001, Florence, Italy.
  Association for Computational Linguistics.

\bibitem[{Rahimi et~al.(2019)Rahimi, Li, and Cohn}]{rahimi-etal-2019-massively}
Afshin Rahimi, Yuan Li, and Trevor Cohn. 2019.
\newblock \href {https://doi.org/10.18653/v1/P19-1015} {Massively multilingual
  transfer for {NER}}.
\newblock In \emph{Proceedings of the 57th Annual Meeting of the Association
  for Computational Linguistics}, pages 151--164, Florence, Italy. Association
  for Computational Linguistics.

\bibitem[{Reimers and Gurevych(2020)}]{reimers2020making}
Nils Reimers and Iryna Gurevych. 2020.
\newblock Making monolingual sentence embeddings multilingual using knowledge
  distillation.
\newblock In \emph{Proceedings of the 2020 Conference on Empirical Methods in
  Natural Language Processing (EMNLP)}, pages 4512--4525.

\bibitem[{Sabou et~al.(2012)Sabou, Bontcheva, and
  Scharl}]{10.1145/2362456.2362479}
Marta Sabou, Kalina Bontcheva, and Arno Scharl. 2012.
\newblock \href {https://doi.org/10.1145/2362456.2362479} {Crowdsourcing
  research opportunities: Lessons from natural language processing}.
\newblock In \emph{Proceedings of the 12th International Conference on
  Knowledge Management and Knowledge Technologies}, i-KNOW '12, New York, NY,
  USA. Association for Computing Machinery.

\bibitem[{Sener and Savarese(2018)}]{sener2018active}
Ozan Sener and Silvio Savarese. 2018.
\newblock Active learning for convolutional neural networks: A core-set
  approach.
\newblock In \emph{International Conference on Learning Representations}.

\bibitem[{Settles and Craven(2008)}]{settles2008analysis}
Burr Settles and Mark Craven. 2008.
\newblock An analysis of active learning strategies for sequence labeling
  tasks.
\newblock In \emph{Proceedings of the 2008 Conference on Empirical Methods in
  Natural Language Processing}, pages 1070--1079.

\bibitem[{S{\o}gaard(2011)}]{sogaard2011data}
Anders S{\o}gaard. 2011.
\newblock Data point selection for cross-language adaptation of dependency
  parsers.
\newblock In \emph{Proceedings of the 49th Annual Meeting of the Association
  for Computational Linguistics: Human Language Technologies}, pages 682--686.

\bibitem[{Song et~al.(2012)Song, Klassen, Xia, and
  Kit}]{song-etal-2012-entropy}
Yan Song, Prescott Klassen, Fei Xia, and Chunyu Kit. 2012.
\newblock \href {https://www.aclweb.org/anthology/C12-2116} {Entropy-based
  training data selection for domain adaptation}.
\newblock In \emph{Proceedings of {COLING} 2012: Posters}, pages 1191--1200,
  Mumbai, India. The COLING 2012 Organizing Committee.

\bibitem[{Stolcke(2002)}]{conf/interspeech/Stolcke02}
Andreas Stolcke. 2002.
\newblock \href
  {http://dblp.uni-trier.de/db/conf/interspeech/interspeech2002.html#Stolcke02}
  {Srilm - an extensible language modeling toolkit.}
\newblock In \emph{INTERSPEECH}. ISCA.

\bibitem[{Tseng et~al.(2020)Tseng, Lee, Huang, and Yang}]{tseng2020cross}
Hung-Yu Tseng, Hsin-Ying Lee, Jia-Bin Huang, and Ming-Hsuan Yang. 2020.
\newblock Cross-domain few-shot classification via learned feature-wise
  transformation.
\newblock \emph{arXiv preprint arXiv:2001.08735}.

\bibitem[{Van~der Maaten and Hinton(2008)}]{van2008visualizing}
Laurens Van~der Maaten and Geoffrey Hinton. 2008.
\newblock Visualizing data using t-sne.
\newblock \emph{Journal of machine learning research}, 9(11).

\bibitem[{Wu and Dredze(2019)}]{wu-dredze-2019-beto}
Shijie Wu and Mark Dredze. 2019.
\newblock \href {https://doi.org/10.18653/v1/D19-1077} {Beto, bentz, becas: The
  surprising cross-lingual effectiveness of {BERT}}.
\newblock In \emph{Proceedings of the 2019 Conference on Empirical Methods in
  Natural Language Processing and the 9th International Joint Conference on
  Natural Language Processing (EMNLP-IJCNLP)}, pages 833--844, Hong Kong,
  China. Association for Computational Linguistics.

\bibitem[{Xue et~al.(2020)Xue, Constant, Roberts, Kale, Al-Rfou, Siddhant,
  Barua, and Raffel}]{xue2020mt5}
Linting Xue, Noah Constant, Adam Roberts, Mihir Kale, Rami Al-Rfou, Aditya
  Siddhant, Aditya Barua, and Colin Raffel. 2020.
\newblock mt5: A massively multilingual pre-trained text-to-text transformer.
\newblock \emph{arXiv preprint arXiv:2010.11934}.

\bibitem[{Yoo and Kweon(2019)}]{yoo2019learning}
Donggeun Yoo and In~So Kweon. 2019.
\newblock Learning loss for active learning.
\newblock In \emph{Proceedings of the IEEE/CVF Conference on Computer Vision
  and Pattern Recognition}, pages 93--102.

\end{thebibliography}

\newpage
\appendix

\section{Data Statistics}
\label{sec:data}
We report the number of sentences in both training and test data in Table \ref{tbl:data1} and \ref{tbl:data2}. POS task lower number of training data relative to NER task for most of the languages. XNLI task has enormous amount of training data compared to POS and NER.
 
 \begin{table}[H]
\centering
\small
\setlength\tabcolsep{2pt}%
\begin{tabular}{*{200}{c}}
\toprule
& \multicolumn{3}{c}{POS}  & \multicolumn{3}{c}{NER} \\
    \cmidrule(lr){2-4}\cmidrule(lr){5-7}
\textbf{Language }& Train & Test & \thead{Language\\ Group} & Train & Test & \thead{Language\\ Group} \\ 
  \midrule
{\small EN} & 11732 & 15039 & - & 19632 & 10000 & -\\
{\small BG} & 8736 & 1116  & $C_1$ & 16235 & 10000 & $C_1$\\
{\small DE} & 149249 & 56354 & $C_1$ & 18515 & 10000 & $C_1$\\
{\small ES} & 14092 & 1278 & $C_1$ & 17817 & 10000 & $C_1$\\
{\small FI} & 14979 & 8233 & $C_1$ & 18933 & 10000 & $C_1$\\
{\small FR} & - & - & - & 18109 & 10000 & $C_1$\\
{\small SV} & 3167 & 3000 & $C_1$ & 14495 & 10000 & $C_1$\\
{\small TR} & 7745 & 9619 & $C_1$ & 18433 & 10000 & $C_1$\\
{\small EL} & 1637 & 456 & $C_1$ & 15908 & 10000 & $C_1$\\
{\small EU} & 5383 & 1799 & $C_2$ & 8089 & 10000 & $C_1$\\
{\small HI} & 13291 & 2000 & $C_2$ & 3948 & 1000 & $C_1$\\
{\small KO} & 22947 & 5563 & $C_2$ & 18796 & 10000 & $C_2$\\
{\small RU} & 3841 & 3601 & $C_1$  & 18795 & 10000 & $C_1$\\
{\small VI} & - & - & - & 16066 & 10000 & $C_1$\\
{\small AR} & 5956 & 2040 & $C_2$  & 17703 & 10000 & $C_2$\\
{\small HE} & 5174 & 491 & $C_2$ & 18329 & 10000 & $C_2$\\
{\small JA} & 7025 & 2172 & $C_2$ & 19012 & 10000 & $C_2$\\
{\small UR} & 3892 & 535 & $C_2$ & 13043 & 1000 & $C_2$\\
{\small ZH} & 3995 & 2451 & $C_2$ & 18310 & 10000 & $C_2$\\
{\small TH} & - & - & - & 17683 & 10000 & $C_3$\\ 
 \bottomrule 
\end{tabular}
\caption {We report the statistics of training and test data available in each language for our experiments.}
  \label{tbl:data1}
 \end{table}

 \begin{table}[H]
\centering
\setlength\tabcolsep{5pt}%
\begin{tabular}{*{200}{c}}
\toprule
\textbf{Language }& Train & Test & \thead{XLMR\\ Group} & \thead{mBERT\\ Group}\\ 
  \midrule
{\small AR} & 392403 & 5010 & $C_1$ & $C_1$ \\
{\small BG} & 392335 & 5010 & $C_1$ & $C_1$ \\
{\small DE} & 392440 & 5010 & $C_1$ & $C_1$ \\
{\small EL} & 392331 & 5010 & $C_1$ & $C_1$ \\
{\small EN} & 392568 & 5010  & $C_1$ & $C_1$\\
{\small ES} & 392405 & 5010 & $C_1$ & $C_1$ \\
{\small FR} & 392405 & 5010 & $C_1$ & $C_1$ \\
{\small HI} & 392356 & 5010  & $C_1$ & $C_2$\\
{\small RU} & 392318 & 5010  & $C_1$ & $C_1$\\
{\small SW} & 391819 & 5010  & $C_1$ & $C_2$\\
{\small TH} & 392480 & 5010 & $C_1$ & $C_2$ \\
{\small TR} & 392177 & 5010 & $C_1$ & $C_1$ \\
{\small UR} & 388826 & 5010 & $C_1$ & $C_2$ \\
{\small VI} & 392416 & 5010 & $C_1$ & $C_1$ \\
{\small ZH} & 392251 & 5010 & $C_1$ & $C_1$ \\
 \bottomrule 
\end{tabular}
\caption {We report the statistics of training and test data available in each language for XNLI.}
  \label{tbl:data2}
 \end{table}

 \section{Study of $\lambda$ and $\gamma$ parameters on few-shot transfer}
 \label{sec:ablation}
 We conduct experiments using the following set of values for $\lambda \in \{0, 0.5, 1\}$ for NER and POS tasks. We have reported the results in Table \ref{tbl:few_shot_ner_en_full} and \ref{tbl:few_shot_pos_en_full}. We find the parameter $\lambda=1$ to be providing highest performance on average for PE, while $\lambda=0.5$ show better performance for $C_2$ language group when $k=10$. For LE methods, we observe $\lambda=0.5$ provides the highest gains in POS tasks. However for NER task, LE method any $\lambda$ parameter provides best gains on average. The gains start diminishing with higher $\lambda$ values in general, but for $C_2$ language group, $\lambda=0.5$ provides best gains for smaller values of $k$.

In Table \ref{tbl:few_shot_xnli_en_full}, we observe that increasing the value $\gamma$ beyond 1 hurts the performance for mBERT. $\gamma=3$ provides higher gains in few cases for XLM-R. But overall, we consider $\gamma=1$ to provide consistent gains across models.

%  For {\small TH}, we find the parameters ($\lambda=0$, $\gamma=0.5$) with highest performance. On POS task, we find the parameters ($\lambda=1$, $\gamma=0.5$)  to be providing highest performance for $C_1$ and $C_2$ language groups. The parameters ($\lambda=0$, $\gamma=0.5$) show highest performance for $C_3$ language group.
 
%  \begin{table}[ht]
% \centering
% \small
% \setlength\tabcolsep{6pt}%
% \begin{tabular}{c|cc}
% \toprule
% Language &  RAND & LE \\ 
% \midrule
% ar & 30.15\% & \textbf{35.21\%} 
% \\
% eu & 16.53\% & \textbf{20.17\%}
% \\
% he & 34.11\% & \textbf{34.77\%} 
% \\
% hi & 28.25\% & \textbf{31.70\%} 
% \\
% ko & 31.65\% & \textbf{34.64\%} 
% \\
% ja & 86.34\% &  \textbf{86.44\%} 
% \\
% ur & \textbf{20.29\%} & 20.26\% 
% \\
% zh & 73.27 & \textbf{77.52} 
% \\
% \bottomrule
% \end{tabular}
% \caption {Percentage of tokens from sentences sampled using {\small RAND} and LE($\lambda=0.5$) for $k=10$ from POS task. We calculated the percentage of tokens from class labels that are mispredicted using the zero-shot model for more than 40\% on the whole language corpus. We observe that the LE method selects sentences containing more incorrect class labels without access to the ground truth labels. The languages from $C_1$ group are not considered as the gains are not relatively low.}
% \label{tbl:misrate_pos}
% % \vspace{-0.5em}
% \end{table}

 \section{Qualitative Analysis of Samples}
 \label{sec:qualitative}
 We have compared sentences selected using {\small RAND} and LE for POS task in Table \ref{tbl:examples_pos}. The comparison of examples from XNLI task selected using {\small RAND} and GE is shown in Table \ref{tbl:examples_xnli}.
 
\begin{table*}[ht]
\centering
\scriptsize
\setlength\tabcolsep{2pt}%
\begin{tabular}{c|p{0.98\linewidth}}
\toprule
 &  Translated Text \\ 
\midrule
 & Study: \hl{Domestic} violence in the \hl{United} States affects 25\% of women and 7.5 of men
\\
\multirow{-2}{*}{\rotatebox[origin=c]{90}{RAND}} & The \hl{Lebanese} authorities, led by the Command of the Emergency Force, contacted the Command and \hl{asked} \hl{them} to move \hl{to prevent} and prevent the violations of Israel, \hl{which} is carrying out works inside the \hl{Lebanese} territories, including especially unloading sand or paving slopes, according to \hl{what} was announced by an \hl{official} source.
\\
\midrule
& It was also agreed to participate in the \hl{Maqama} \hl{cultural} festivals held on both sides by the two sides and to exchange and exchange musical, folk bands, \hl{artists}, \hl{painters}, \hl{playwrights}, and \hl{others}.
\\
 \multirow{-5}{*}{\rotatebox[origin=c]{90}{LE}} & \hl{Arafat} is \hl{also} \hl{scheduled} to visit \hl{Kuala} \hl{Lumpur}, \hl{Jakarta}, and \hl{Tokyo}, \hl{according} to \hl{Palestinian} sources.
 \\
\bottomrule
\end{tabular}
\caption {Sample sentences in {\small AR} from POS task using {\small RAND} and LE ($\lambda=0.5$) methods for $k=10$ using XLM-R. The tokens are highlighted having ground truth class labels that are mispredicted using the zero-shot model. In case for {\small AR}, we noticed following class labeled are wrongly predicted: Other, Interjection, Particle, Adjective, Determiner, Pronoun and Adverb. LE ($\gamma=0.5$) select sentences containing these class labels more frequently than {\small RAND}.}
\label{tbl:examples_pos}
% \vspace{-0.5em}
\end{table*}

\begin{table*}[ht]
\centering
\scriptsize
\setlength\tabcolsep{2pt}%
\begin{tabular}{c|p{0.48\linewidth}|p{0.48\linewidth}}
\toprule
 &  Translated Premise & Translated Hypothesis \\ 
\midrule
 & The 2000 census is the most important and provides valuable information. & Census Monitoring Board to oversee the 2000 decennial census. 
\\
 & The guarantee is a really poor tree guarantee. & Yes I know the guy at wolfe told us they cut the tree warranty like six months or less.
\\
& This drama isn't as interesting as nice celebrity meat, according to all the major gossip sites. & Who cares what David Hare did to Arthur Schnitzler's la ronde when there's celebrity meat to be suffered?
 \\
& I am a californian. & Oh yes yes yes i'm from i'm from up around i'm a new yorker myself.
\\
 & They would try to kill him like a pack of savages. & They would come down on him like a pack of blood thirsty wolves. 
\\
\multirow{-7}{*}{\rotatebox[origin=c]{90}{RAND}} & Egyptians travel to surrounding countries to visit casinos. & Egyptian nationals are not allowed to gamble so casinos are only open to foreign guests over the age of 21 (you will be asked for ID).
\\
\midrule
 & People from France are usually very boring. &  For a people so proud of their identity, the French are a rich mix. 
\\
 & The French are proud, dynamic patriots. & For a people so proud of their identity, the French are a rich mix. 
\\
 & They like to sit in fancy places. & 'Cause they're fancy places and stuff.
\\
 & That's because they're fancy seats and all. & 'Cause they're fancy places and stuff.
\\
 & If they apologize for forgetting their name, the awkwawrdness will all go away. & If nothing is done to alleviate the situation , you can say bluntly , I 'm so sorry I 've forgotten her name.
\\
 \multirow{-7}{*}{\rotatebox[origin=c]{90}{GE}} & If you can't do better, just apologize for forgetting her name. & If nothing is done to alleviate the situation, you can say, bluntly, I'm so sorry I can't remember her name.
 \\
\bottomrule
\end{tabular}
\caption {Sample sentences in {\small DE} from XNLI task using {\small RAND} and GE ($\gamma=1$) methods for $k=10$ using XLM-R. We observe that GE ($\gamma=1$) select two examples having similar context but different labels.}
\label{tbl:examples_xnli}
% \vspace{-0.5em}
\end{table*}

\begin{table*}[h]
\centering
\small
\setlength\tabcolsep{5.5pt}%
\begin{tabular}{c|c|cccccc|cccccc}
\toprule
\multicolumn{1}{c}{} & & \multicolumn{6}{c}{$S=400k$}  & \multicolumn{6}{c}{$S=40k$} \\
    \cmidrule(lr){3-8} \cmidrule(lr){9-14}
\multicolumn{1}{c}{}  & \textbf{Method} & $10$ & $100$ & $500$ & $1k$ & $5k$ & $10k$ & $10$ & $100$ & $500$ & $1k$ & $5k$ & $10k$\\ 
\midrule
 & RAND & -0.5 &
-0.9 &
-0.2 &
0.6 &
2.4 &
3.4 &
-1.2 &
-0.7 &
0.0 &
0.9 &
2.8 &
3.8
 \\ 
 & DCE & 0.0 &
-0.1 &
-0.1 &
0.5 &
2.8 &
3.9 &
0.2 &
-0.3 &
0.2 &
0.7 &
3.6 &
4.3
 \\
 & PE ($\lambda=1$) &   
 -0.4 &
-1.0 &
-0.3 &
0.4 &
2.6 &
3.4 &
- &
- &
- &
- &
- &
-
\\
 & GE & -0.1 &
-0.9 &
-0.5 &
0.2 &
2.6 &
3.1 & - &
- &
- &
- &
- &
-
\\
 & GE ($\gamma =1$) & 
\textbf{0.1} &
\textbf{-0.1} &
\textbf{0.7} &
\textbf{1.4} &
\textbf{3.3} &
\textbf{4.2} &
\textbf{0.3} &
\textbf{0.7} &
\textbf{0.6} &
\textbf{1.9} &
\textbf{3.9} &
\textbf{4.6}
 \\ 
 & GE ($\gamma =2$) &
-0.3 &
-0.7 &
-0.1 &
1.3 &
3.1 &
4.1 & - &
- &
- &
- &
- &
-
\\
\multirow{-6}{*}{\rotatebox[origin=c]{91}{mBERT}} & GE ($\gamma =3$) &
-0.6 &
-0.2 &
0.2 &
1.3 &
\textbf{3.3} &
4.1 & - &
- &
- &
- &
- &
-
\\
\midrule
 & RAND & -0.1 &
-0.2 &
0.0 &
0.2 &
1.2 &
1.5 &
-0.7 &
-0.6 &
-0.6 &
0.1 &
1.1 &
1.5
 \\ 
 & DCE & {0.2} &
\textbf{0.7} &
0.5 &
0.7 &
1.6 &
{2.1} &
-0.4 &
-0.9 &
-0.2 &
0.1 &
1.5 &
1.9
 \\
 & PE ($\lambda=1$) 
 -0.4 &
-0.0 &
0.0 &
0.3 &
1.3 &
1.5 &
- &
- &
- &
- &
- &
-
\\
  & GE & -0.3 &
-0.1 &
0.2 &
0.4 &
1.5 &
\textbf{2.6} & - &
- &
- &
- &
- &
-
\\
& GE ($\gamma =1$) &
0.1 &
0.5 &
{0.6} &
\textbf{1.2} &
{1.9} &
{2.1} &
\textbf{-0.1} &
\textbf{0.5} &
\textbf{-0.1} &
\textbf{0.6} &
\textbf{1.6} &
\textbf{2.0}
 \\ 
 & GE ($\gamma =2$) &
 0.4 &
0.3 &
0.7 &
1.3 &
2.1 &
2.3 &
- &
- &
- &
- &
- &
-
 \\
 \multirow{-6}{*}{\rotatebox[origin=c]{91}{mBERT}} & GE ($\gamma =3$)  & \textbf{0.7} &
0.3 &
\textbf{1.1} &
1.0 &
\textbf{2.1} &
2.2 &
- &
- &
- &
- &
- &
-
 \\
\bottomrule
\end{tabular} 
\caption {Few-shot performance on XNLI tasks with varying number of target language examples $k$ using {\small EN} as the pivot language. We have reported the $\triangle$ delta between few-shot and zero-shot performance averaged across all languages. $S$ denotes the size of the pivot-language corpus.}
\label{tbl:few_shot_xnli_en_full}
% \vspace{-0.5em}
\end{table*}

\begin{table*}[ht]
\centering
\small
\setlength\tabcolsep{4pt}%
\begin{tabular}{c|c|ccc|ccc|ccc|ccc|ccc}
\toprule
% & $k$  & 10  & 50 & 100 & 500 & 1000 \\
\multicolumn{1}{c}{} & & \multicolumn{3}{c}{$k=10$}  & \multicolumn{3}{|c}{$k=50$} & \multicolumn{3}{|c}{$k=100$} & \multicolumn{3}{|c}{$k=500$} & \multicolumn{3}{|c}{$k=1000$} \\
    \cmidrule(lr){3-5}\cmidrule(lr){6-8}\cmidrule(lr){9-11}  \cmidrule(lr){12-14} \cmidrule(lr){15-17} 
\multicolumn{1}{c}{} & \multirow{-1}{*}{\textbf{Method}} & $\triangle^{C_1}$   & $\triangle^{C_2}$ &  $\triangle^{C_3}$ & $\triangle^{C_1}$  & $\triangle^{C_2}$ &  $\triangle^{C_3}$ & $\triangle^{C_1}$  & $\triangle^{C_2}$ &  $\triangle^{C_3}$ & $\triangle^{C_1}$  & $\triangle^{C_2}$ &  $\triangle^{C_3}$ & $\triangle^{C_1}$  & $\triangle^{C_2}$ &  $\triangle^{C_3}$ \\ 
\midrule
& RAND & 2.9 & 9.9 & 0.5 &
6.4 & 15.8 & 1.3 &
7.7 & 17.4 & 1.3 &
12.1 & 26.9 & 18.6 &
14.0 & 30.4 & {31.2}
\\ 
& DCE & 1.8 & 8.4 & {4.0} &
5.1 & 12.8 & {2.8} &
5.2 & 11.8 & {3.3} &
9.9 & 23.0 & 18.8 &
12.3 & 28.5 & 29.2
\\
 & PE ($\lambda = 0$) &
 3.9 & 12.0 & 1.6 &
6.3 & 16.7 & 0.4 &
7.6 & 17.8 & 0.3 &
12.2 & 27.1 & 18.3 &
13.7 & 30.0 & \textbf{32.5}
 \\
 & PE ($\lambda = 0.5$) &
 4.4 & {13.2} & 0.2 &
7.6 & 18.1 & 1.0 &
8.2 & 18.8 & 0.0 &
12.5 & \textbf{27.8} & 18.5 &
14.2 & 30.3 & 32.1
 \\
 & PE ($\lambda = 1$) & 3.1 & 10.0 & \textbf{5.7} &
6.8 & 14.5 & \textbf{4.7} &
7.5 & 16.1 & \textbf{3.9} &
12.4 & 24.5 & \textbf{19.8} &
14.2 & 27.4 & 27.4
\\
& LE & 5.5 & {11.3} & 2.5 &
7.4 & \textbf{18.4} & 1.1 &
\textbf{8.9} & {18.9} & 0.7 &
\textbf{13.0} & {27.6} & {18.9} &
{14.9} & \textbf{30.6} & 31.0
\\
 & LE ($\lambda=0$) & \textbf{5.6} & 11.0 & 0.7 &
\textbf{8.4} & 18.3 & 0.1 &
8.7 & 18.4 & -0.0 &
12.9 & 26.9 & 15.1 &
14.8 & 30.0 & 29.8
\\
\multirow{-8}{*}{\rotatebox[origin=c]{90}{mBERT}} & LE ($\lambda=0.5$) & 
4.4 & \textbf{13.9} & 0.4 &
7.6 & 17.3 & -0.1 &
8.6 & \textbf{19.1} & -0.4 &
12.7 & 27.5 & 12.9 &
\textbf{15.0} & 30.0 & 27.9
\\
\midrule
& RAND & 1.4 & 8.0 & 0.3 &
6.7 & 15.3 & 0.4 &
7.8 & 16.8 & 1.5 &
12.8 & \textbf{26.1} & \textbf{20.7} &
\textbf{14.6} & {29.4} & {27.7}
\\ 
& DCE & -3.8 & 0.5 & 2.4 &
3.4 & 10.0 & 0.9 &
4.3 & 10.5 & -0.2 &
10.5 & 23.8 & 19.2 &
13.2 & 27.8 & 26.3
\\
 & PE ($\lambda = 0$) & 1.1 & 8.5 & 1.5 &
6.2 & 14.7 & \textbf{7.3} &
7.8 & 15.6 & 6.4 &
\textbf{13.0} & 25.3 & 22.1 &
14.7 & 29.0 & 28.1
\\
 & PE ($\lambda = 0.5$) &
 2.5 & {10.9} & -1.2 &
7.6 & 15.4 & 2.3 &
8.6 & 17.0 & -1.0 &
\textbf{13.0} & 25.4 & 20.1 &
14.9 & 28.7 & \textbf{28.8}
\\
 & PE ($\lambda = 1$) & \textbf{4.4} & {8.6} & {3.6} &
7.0 & {16.5} & 1.2 &
7.9 & \textbf{17.8} & 0.5 &
12.3 & \textbf{26.1} & 19.0 &
14.2 & \textbf{29.9} & {28.4} 
\\
& LE & {3.0} & 8.1 & \textbf{5.7} &
\textbf{7.9} & 15.7 & 3.4 &
\textbf{9.0} & 16.7 & 2.4 &
\textbf{13.0} & \textbf{26.1} & 18.2 &
14.5 & 29.1 & 23.4
\\
& LE ($\lambda=0$) & 2.4 & {8.2} & 1.4 &
7.4 & {16.0} & {5.0} &
8.5 & {16.9} & \textbf{4.0} &
\textbf{13.0} & \textbf{26.1} & 16.5 &
14.5 & 29.3 & 23.2
\\
\multirow{-8}{*}{\rotatebox[origin=c]{90}{XLM-R}} & LE ($\lambda=0.5$) & 
2.4 & \textbf{11.0} & 2.8 &
7.6 & \textbf{16.6} & 5.2 &
8.8 & 16.0 & 2.5 &
12.6 & 25.9 & 16.3 &
14.4 & 28.6 & 22.8
\\
\bottomrule
\end{tabular}
\caption {Few-shot cross-lingual transfer performance on NER tasks with varying number of target language examples $k$ using {\small EN} as the pivot language. We have reported the $\triangle$ delta between few-shot and zero-shot performance averaged across the languages in each category $C_1$ and $C_2$, and $C_3$.} 
\label{tbl:few_shot_ner_en_full}
% \vspace{-0.5em}
\end{table*}

\begin{table*}[ht]
\centering
\small
\setlength\tabcolsep{6pt}%
\begin{tabular}{c|c|cc|cc|cc|cc|cc}
\toprule
% & $k$  & 10  & 50 & 100 & 500 & 1000 \\
\multicolumn{1}{c}{} & & \multicolumn{2}{c}{$k=10$}  & \multicolumn{2}{|c}{$k=50$} & \multicolumn{2}{|c}{$k=100$} & \multicolumn{2}{|c}{$k=500$} & \multicolumn{2}{|c}{$k=1000$}
\\
    \cmidrule(lr){3-4}\cmidrule(lr){5-6}\cmidrule(lr){7-8} 
    \cmidrule(lr){9-10} \cmidrule(lr){11-12}
\multicolumn{1}{c}{} & \multirow{-1}{*}{\textbf{Method}} & $\triangle^{C_1}$   & $\triangle^{C_2}$ & $\triangle^{C_1}$  & $\triangle^{C_2}$ & $\triangle^{C_1}$  & $\triangle^{C_2}$  
& $\triangle^{C_1}$  & $\triangle^{C_2}$ & $\triangle^{C_1}$  & $\triangle^{C_2}$
\\ 
\midrule
 & Rand & 4.1 & 22.6 &
6.7 & 27.5 &
7.3 & 28.0 &
11.9 & 46.5 &
12.5 & 48.4
\\
 & DCE & 2.3 & 18.7 &
5.2 & 24.3 &
6.0 & 25.9 &
11.8 & 46.1 &
12.4 & 48.3
\\
& PE ($\lambda = 0$) & 
4.9 & 22.5 &
6.9 & 27.7 &
7.4 & 28.1 &
10.2 & 38.8 &
10.9 & 40.5
\\
& PE ($\lambda = 0.5$) & 
\textbf{5.2} & 22.5 &
7.0 & 28.0 &
\textbf{7.5} & 28.3 &
10.4 & 38.9 &
11.2 & 40.5
\\
 & PE ($\lambda = 1$) & 4.4 & \textbf{23.4} &
\textbf{7.0} & 27.8 &
7.4 & 28.1 &
10.5 & 38.8 &
11.2 & 40.4
\\
& LE & 3.9 & 20.1 &
6.3 & 26.3 &
7.1 & 26.9 &
\textbf{12.1} & 46.9 &
\textbf{12.8} & {48.7}
\\
& LE ($\lambda=0$) & {4.5} & 21.9 &
{6.8} & 27.3 &
\textbf{7.5} & 27.9 &
\textbf{12.1} & \textbf{47.0} &
12.0 & {48.7}
\\
\multirow{-8}{*}{\rotatebox[origin=c]{90}{mBERT}} & LE ($\lambda=0.5$) & 4.1 & {23.3} &
{6.8} & \textbf{28.2} &
\textbf{7.5} & \textbf{28.6} &
11.3 & \textbf{47.0} &
11.5 & \textbf{48.8}
\\
\midrule
 & RAND & 3.1 & 24.6 &
5.2 & 28.5 &
5.6 & 28.8 &
9.5 & 42.3 &
9.8 & 44.6
\\
 & DCE & 1.8 & 22.1 &
3.7 & 26.0 &
4.5 & 27.2 &
9.4 & 42.0 &
9.8 & 44.5
\\
& PE ($\lambda = 0$) & 
4.0 & 24.7 &
5.6 & 28.6 &
5.9 & 29.0 &
8.5 & 39.4 &
8.9 & 41.3
\\
& PE ($\lambda = 0.5$) & 
\textbf{4.1} & 25.1 &
\textbf{5.9} & 28.8 &
6.2 & \textbf{29.2} &
8.8 & 39.4 &
9.1 & 41.2
\\
 & PE ($\lambda = 1$) & 3.4 & 24.7 &
5.6 & 29.1 &
6.1 & \textbf{29.2} &
8.8 & 39.2 &
9.1 & 41.1
\\
& LE & 2.9 & 22.1 &
5.5 & 27.6 &
6.2 & 28.5 &
\textbf{9.6 } & 42.6 &
\textbf{9.9} & \textbf{44.9}
\\
& LE ($\lambda=0$) & 3.1 & 24.9 &
5.6 & 28.6 &
6.1 & 28.8 &
9.5 & 42.7 & 
9.3 & \textbf{44.9}
\\
\multirow{-8}{*}{\rotatebox[origin=c]{90}{XLM-R}} & LE ($\lambda=0.5$) & {3.5} & \textbf{25.2} &
\textbf{5.8} & \textbf{29.2} &
\textbf{6.4} & \textbf{29.2} &
9.0 & \textbf{42.8} &
8.9 & \textbf{44.9}
\\
\bottomrule
\end{tabular}
\caption {Few-shot cross-lingual transfer performance on POS tasks with varying number of target language examples $k$ using {\small EN} as the pivot language. We have reported the $\triangle$ delta between few-shot and zero-shot performance averaged across the languages in each category $C_1$ and $C_2$.} 
\label{tbl:few_shot_pos_en_full}
% \vspace{-0.5em}
\end{table*}

\end{document}